\documentclass[10pt]{article}
\usepackage{amsmath}
\usepackage{amsfonts}
\usepackage{latexsym}
\usepackage{graphicx}
\usepackage{amssymb}
\usepackage{mathabx}   
\newcommand{\R}{\mathbb{R}}

\begin{document}

\title{Ultrametric Component Analysis with Application to Analysis of 
Text and of Emotion}
\author{Fionn Murtagh \\
School of Computer Science and Informatics \\
De Montfort University, Leicester LE1 9BH, England \\
fmurtagh@acm.org}
\maketitle

\begin{abstract}
We review the theory and practice of determining what parts of a data set
are ultrametric.  It is assumed that the data set, to begin with, is endowed
with a metric, and we include discussion of how this can be brought about 
if a dissimilarity, only, holds.  The basis for part of the metric-endowed data 
set being ultrametric is to consider triplets of the observables (vectors).   
We develop a novel consensus of hierarchical clusterings.  We do this in order
to have a framework (including visualization and supporting interpretation) 
for the parts of the data that are determined to be ultrametric.  Furthermore 
a major objective is to determine locally ultrametric relationships as opposed
to non-local ultrametric relationships.  As part of 
this work, we also study a particular property of our ultrametricity coefficient, 
namely, it being a function of the difference of angles of the base angles of
the isosceles triangle.  This work is completed by a review of related work,
on consensus hierarchies, and of a major new application, namely 
quantifying and interpreting the emotional content of narrative.   
\end{abstract}

Topics: 

\begin{enumerate}

\item Introduction
\begin{enumerate}
\item Metric and Ultrametric
\item Transforming to a Metric
\item Transforming to an Ultrametric
\begin{enumerate}
\item Agglomerative Hierarchical Clustering Algorithms
\item Hierarchical Clustering and Associated Ultrametric
\item Agglomerative Criterion and Its Associated Ultrametric
\end{enumerate}
\end{enumerate}

\item Quantifying How Metric or How Ultrametric A Data Set Is
\begin{enumerate}
\item Quantifying How Metric a Data Set Is
\item Quantifying How Ultrametric a Data Set Is
\begin{enumerate}
\item Ultrametricity Coefficient of Lerman
\item Ultrametricity Coefficient of Rammal, Toulouse and Virasoro
\item Ultrametricity Coefficients of Treves and of Hartmann
\item Bayesian Network Modeling
\item Our $\alpha_\epsilon$-Ultrametricity Coefficient
\item Application to Partial Ultrametric Embedding
\end{enumerate}
\end{enumerate}

\item Transforming Data to Become More Metric or More Ultrametric
\begin{enumerate}
\item Transforming Non-Metric to Metric Data
\item Transforming Metric, or Non-Metric, to Minimal Superior
Ultrametric, and to Maximal Inferior Ultrametric
\item Approximating an Ultrametric for Similarity Metric
Space Searching
\end{enumerate}

\item Emotion as the Doorway into the Subconscious
\begin{enumerate}
\item Emotion as an Essential Component of Perception
\item Data With Relatively High Ultrametricity
\item A New Principle for Metric and Ultrametric Component Analysis
\end{enumerate}
\item Consensus Ultrametric Sets from Two Hierarchies
\begin{enumerate}
\item Ultrametric Consensus: Definition and Algorithm
\item Application: Ultrametric Consensus of Hierarchies
\item Inversions in Hierarchies Clustering
\item Ultrametric Consensus 
\item From the Ultrametric Consensus Set to a Consensus Hierarchy
\end{enumerate}
\item Taking the Ultrametric Consensus Set Beyond Model Fitting:
Inherent Ultrametric Properties
\begin{enumerate}
\item Coming from a Space Endowed at least with a Dissimilarity, 
and Analyzing it when Mapped into an Ultrametric Space
\item Ultrametric Triplet Sets Through Other Pairs of Hierarchies
\item Just How Good Are the Ultrametric-Respecting Triplets                
Relative to the Input Data?
\item Interpretation of Ultrametric Triplets
\end{enumerate}
\item Conclusions 
\item Appendix  
\begin{enumerate}
\item A1. Comparing and Combining Hierarchical Clusterings
\item A2. Metrics and Ultrametrics based on 3-Way Distances or Dissimilarities
\item A3. Ultrametric Alpha Epsilon-Respecting Triplets Using
Ward and Single Link
\end{enumerate}
\item References
\end{enumerate}

\section{Introduction}

%
%
%
%

\subsection{Metric and Ultrametric}

A metric space $(X,d)$ consists of a set~$X$ on which is defined a
distance function $d$ which assigns to each pair of points of
$X$ a distance between them, and satisfies the following four axioms
for any triplet of points $x, y , z$:

\begin{equation*}
\mbox{A1: } \forall x, y \in X,  d(x,y) \geq 0 \mbox{ (positiveness)}
\end{equation*}

\begin{equation*}
\mbox{A2: } \forall x, y \in X, d(x,y) = 0 \mbox{ iff } x = y \mbox{ (reflexivity)}
\end{equation*}

\begin{equation*}
\mbox{A3: } \forall x, y \in X, d(x,y) = d(y,x)  \mbox{ (symmetry)}
\end{equation*}

\begin{equation*}
\mbox{A4: } \forall x, y, z \in X, d(x,z) \leq d(x,y) + d(y,z)  \mbox{ 
(triangular inequality)}
\end{equation*}

When considering an ultrametric space we need to consider
the strong triangular inequality or ultrametric inequality defined as:

\begin{equation*}
\mbox{A5: } d(x,z) \leq max~\{d(x,y), \ d(y,z)\} \mbox{ (ultrametric inequality)}
\end{equation*}
and this in addition to the positivity, reflexivity and symmetry properties
(properties A1, A2, A3)
for any triple of point  $x, y, z \in X$.

\subsection{Transforming to a Metric}

Very briefly, we summarize the various mappings that are commonly 
and widely applied.

\begin{enumerate}
\item Starting with observations with readings on attributes, arranged
as rows and columns of a matrix, $X$, then the spectral reduction of 
the variance-covariance matrix, $X^\prime X$, where $X$ may in addition
have preprocessing involving centering of attributes, or centering and 
reduction to unit variance of attributes, yields the principal components, 
i.e.\ an orthonormal coordinate system endowed with the Euclidean
distance.  The input is assumed to be endowed with the Euclidean 
distance also.  The three cases considered (centering, reduction, or
not) provide for PCA of (i) the sums of squares and cross-products 
matrix, (ii) variances-covariances, and (iii) correlations.  

\item When qualitative data are at issue, then Correspondence Analysis is
the more appropriate analysis method, through projection into a Euclidean
factor space.  See \cite{murtagh2000,MurtaghCA}.  For contingency table 
data, a natural model for the data as quantified by the $\chi^2$ statistic
is the degree of fit between observed data and the expected data given 
by the outer product of the marginal, and also the row and column mass,
 densities.   It is this discrepancy from
the model provided by this product of densities that is interpreted in 
the Correspondence Analysis. 

\item Principal Coordinates Analysis takes distances as input and 
reconstructs a coordinate space.   See section \ref{PCoorA}. 
\end{enumerate}

\subsection{Transforming to an Ultrametric}
\label{transforming}

Here we consider the transforming of data endowed with a metric
to the same set endowed with an ultrametric.  We do not require
a metric, and instead of a distance the algorithm performing
the transformation to an ultrametric can work on input consisting
of dissimilarities.  (Recall that a dissimilarity is not required
to satisfy the triangular inequality.)

\subsubsection{Agglomerative Hierarchical Clustering Algorithms}
\label{AHC}

If $X$ is endowed with a metric, then this metric can be
mapped onto an ultrametric.  In practice, endowing  $X$
with a metric can be relaxed to a dissimilarity.  An often
used mapping from metric to ultrametric
is by means of an agglomerative hierarchical clustering
algorithm.  A succession of $n - 1$ pairwise merge steps takes place by making
use of the closest pair of singletons and/or clusters at each step.  Here
$n$ is the number of observations, i.e.\ the cardinality of set $X$.
Closeness between
singletons is furnished by whatever distance or dissimilarity is in use.
For closeness between singleton or non-singleton clusters, we need to
define an inter-cluster distance or dissimilarity.  This can be defined
with reference to the cluster compactness or other property that we wish
to optimize at each step of the algorithm.   

\subsubsection{Hierarchical Clustering and Associated Ultrametric}
\label{sectUMdist}

A hierarchy, $H$,
is defined as a binary, rooted, node-ranked tree, also
termed a dendrogram \cite{benz,john,Lerman81,Murtagh85-1}.
A hierarchy defines a set of embedded subsets of a given set of objects
$X$, indexed by the set $I$.
These subsets are totally ordered by an index function $\nu$, which is a
stronger condition than the partial order required by the subset relation.
A bijection exists between a hierarchy and an ultrametric space.

Let us show these equivalences between embedded subsets, hierarchy, and
binary tree, through the constructive approach of inducing $H$ on a set
$I$.

Hierarchical agglomeration on $n$ observation vectors with indices
$i \in I$ involves
a series of $1, 2, \dots , n-1$ pairwise agglomerations of
observations or clusters, with properties that follow.

In order to
simplify notation, let us use the index $i$ to represent also the
observation, and also the observation vector.  Hence for $i = 3$ and
the third -- in some sequence -- observation vector, $x_i = x_3$, we
will use $i$ to also represent $x_i$ in such a case.

A hierarchy
$H = \{ q | q \in 2^I \} $ such that (i) $I \in H$, (ii) $i \in H \ \forall 
i$, and (iii) for each $q \in H, q^\prime \in H: q \cap q^\prime \neq 
\emptyset \Longrightarrow q \subset  q^\prime \mbox{ or }  q^\prime 
 \subset q$.  Here we have denoted the power set of set $I$ by $2^I$.
An indexed hierarchy is the pair $(H, \nu)$ where the positive
function defined on $H$, i.e., $\nu : H \rightarrow \R^+$, satisfies:
$\nu(i) = 0$ if $i \in H$ is a singleton; and (ii)  $q \subset  q^\prime 
\Longrightarrow \nu(q) < \nu(q^\prime)$.  Here we have denoted the
positive reals, including 0, by $\R^+$.
Function $\nu$ is the agglomeration
level.  Take  
$q \subset q''$
 and $q^\prime \subset q''$, and let $q''$ be the lowest level cluster for
which this is true. Then if we define $D(q, q^\prime) = \nu(q'')$, $D$ is
an ultrametric.

A hierarchical clustering tree is referred to as a dendrogram. 

\subsubsection{Agglomerative Criterion and Its Associated Ultrametric}
\label{msmi}

In practice, we start with a Euclidean or alternative
dissimilarity, use some criterion such as minimizing the change in variance
resulting from the agglomerations, or the minimal linkage (in terms of
the initial input dissimilarities, and the redefined dissimilarities between
clusters as they are formed) and then define $\nu(q)$ as the
dissimilarity associated with the agglomeration carried out.

For observations $i, i' \in I$, let $q$ be the least, with respect to 
set inclusion, cluster containing both $i$ and $i'$: $ i \neq i'; 
i \in q \mbox{ and } i' \in q$.
Then the ultrametric distance between $i$ and $i'$ is 
$D(i,i') = \nu(q)$.    

With the initial dissimilarity or distance, $d(i,i')$, the 
hierarchical agglomerative clustering is specified as follows: 
given a set of clusters comprising a partition $P$ of $I$, and defining a 
level of the hierarchical clustering, then agglomerate $q, q' \in P$ such 
that $d(q,q') \leq d(q'', q'''), \forall q'', q''' \in P$.  Finally 
the agglomerative clustering criterion allows us to define $d$.  

For the single link agglomerative clustering method, $d(q,q') = 
\min_{i,i'} d(i,i'), i \in q, i \in q'$.   Single link is also referred
to as the minimum link or nearest neighbour 
method.  The ultrametric distance, read off the 
dendrogram, $d_u(i,i')$ is such that $d_u(i,i') \leq d(i,i')$ and it is the 
maximum such value over all ultrametric distance matches (i.e., 
mappings) of the 
input dissimilarity or distance data.  For this reason the single
link agglomerative clustering algorithm is referred to as the subdominant
ultrametric, and also the maximum inferior ultrametric.  See 
\cite{benz,jardine}.

Parenthetically we note the close link between the single link method
and the minimal spanning tree.  Kruskal's algorithm \cite{kruskal56}
takes the edges, i.e.\ pairs $(i, i')$  in increasing order of their 
distance (or dissimilarity), $d(i,i')$, and connects them if a graph 
cycle does not result.  In this way we end up with a tree (because cycles
are excluded), that is spanning (all points $i, i'$ are included), and that 
is minimal (the sum of edge weights or dissimilarities $d(i,i')$ over all 
connected pairs, $(i,i')$, is minimal).  Note that the link formed by a pair 
$(i,i')$ is preserved in the minimal spanning tree, whereas the merger of 
two clusters $(q,q')$ in the single link hierarchical clustering does not 
preserve this information; it follows that transforming the minimal spanning 
tree into the single link hierarchical clustering only requires the processing
of the $n - 1$ edges; whereas the transforming of the single link hierarchy 
into the minimal spanning tree requires more processing, determining the 
minimal distance between members of clusters for every pair of clusters 
(immediate if the clusters are singletons).  The latter task is easily 
seen to be quadratic if we approach it as the building from the start of the
single link hierarchical clustering, because in doing so we have the required
(closest link between clusters) information.  

In the case of the complete link method, we are dealing with a 
maximum link method.  The ultrametric distance, read off the
dendrogram, $d_u(i,i')$ is such that $d_u(i,i') \geq d(i,i')$ and it is
a non-unique minimal 
such value over all ultrametric distance matches of the
input dissimilarity or distance data.  For this reason the complete
link agglomerative clustering algorithm is referred to as a 
 minimal superior ultrametric.  See \cite{benz}.  Another name used
is furthest neighbour algorithm.  

\section{Quantifying How Metric or How Ultrametric A Data Set Is}
\label{quantifying}

\subsection{Quantifying How Metric a Data Set Is}
\label{PCoorA}

We will take a practical and applicable framework in order to 
show how the positive (or non-negative) eigenvalues of a spectral 
decomposition provides a way to measure how metric a data set is.  

Consider Principal Coordinates Analysis, also referred to as
Classical Multidimensional Scaling and metric scaling, and 
associated with the names of Torgerson \cite{torgerson} and Gower.  
It takes distances as input and produces coordinate values as 
output. 

Consider the initial data
matrix, $X$, of dimensions $n \times m$, and the ``sums of squares and
cross products'' matrix of the rows:
$$ A = XX^\prime $$
$$ a_{ik} = \sum_j x_{ij} x_{kj}. $$
If $d_{ik}$ is the Euclidean distance between objects $i$ and $k$
(using row vectors $i$ and $k$ of matrix $X$) we have that:
$$ d^2_{ik} = \sum_j (x_{ij} - x_{kj})^2 $$
$$ = \sum_j x^2_{ij} + \sum_j x^2_{kj} - 2 \sum_j x_{ij} x_{kj} $$
\begin{equation}
\setcounter{equation}{1}
 = a_{ii} + a_{kk} - 2 a_{ik} .
\label{eqnddd}
\end{equation}
In Principal Coordinates Analysis, we are given the distances and we
want to obtain $X$.  We will assume that the columns of this matrix
are centered, i.e.
$$ \sum_i x_{ij} = 0 . $$

It will now be shown that matrix $A$ can be constructed from the
distances using the following formula: 
\begin{equation}
 a_{ik} = - {1 \over 2} (d^2_{ik} - d^2_i - d^2_k - d^2)
\end{equation}
where
$$ d^2_i = {1 \over n} \sum_k  d^2_{ik} $$
$$ d^2_k = {1 \over n} \sum_i  d^2_{ik} $$
$$ d^2 = {1 \over n^2} \sum_i \sum_k d^2_{ik} . $$
This result may be proved by substituting for the distance terms
(using equation \ref{eqnddd}), and 
knowing that by virtue of the centering of the
row vectors of matrix $X$, we have
$$ \sum_i a_{ik} = 0$$
(since $a_{ik} = \sum_j x_{ij} x_{kj}$; and consequently in the term
$\sum_i \sum_j x_{ij} x_{kj}$ we can separate out $\sum_i x_{ij}$ which
equals zero).
Similarly (by virtue of the symmetry of $A$) we use the fact that
$$ \sum_k a_{ik} = 0 . $$
Having thus been given distances, we have constructed matrix 
$A = XX^\prime $.  We now wish to reconstruct matrix $X$; or, since this
matrix has in fact never existed, we require some matrix $X$ which
satisfies $XX^\prime = A$.

If matrix $A$ is positive, symmetric and semidefinite, it will have
rank $ p \leq n$.  We may derive $p$ non-zero eigenvalues, 
$\lambda_1 \geq \lambda_2 \geq \dots \geq \lambda_p > 0$, with corresponding
eigenvectors ${\bf u}_1, {\bf u}_2, \dots, {\bf u}_p$.  Consider the 
scaled eigenvectors, defined as ${\bf f}_i = \sqrt{\lambda_i} {\bf u}_i$.
Then the matrix $X = ({\bf f}_1, {\bf f}_2, \dots, {\bf f}_p)$ is a 
possible coordinate matrix.  This is shown as follows.  We have, in 
performing the eigen-decomposition of $A$:
$$ A {\bf u}_i = \lambda_i {\bf u}_i $$
and by requirement
$$ XX^\prime {\bf u}_i = \lambda_i {\bf u}_i . $$
In the left hand side, 
$ X \left( \begin{array}{c}
            {\bf f}_1 \\
            {\bf f}_2 \\
              .   \\  . \\  .  \\
            {\bf f}_p   \end{array}  \right) {\bf u}_i
 =  X \left( \begin{array}{c}
            \sqrt{\lambda_1}{\bf u}_1 \\
            \sqrt{\lambda_2}{\bf u}_2 \\
              .   \\  . \\  .  \\
            \sqrt{\lambda_p}{\bf u}_p  \end{array}   \right){\bf u}_i 
 = X \left( \begin{array}{c}
               0 \\  0 \\ .  \\ \sqrt{\lambda_i} \\  . \\ 0
                      \end{array}      \right) $
since eigenvectors are mutually orthogonal.  Continuing:

$ (\sqrt{\lambda_1}{\bf u}_1, \sqrt{\lambda_2}{\bf u}_2, \dots, 
  \sqrt{\lambda_p}{\bf u}_p)  \left( \begin{array}{c}
0 \\  0 \\ .  \\ \sqrt{\lambda_i} \\  . \\ 0
                  \end{array}   \right) =  \lambda_i {\bf u}_i . $ 

Thus we have succeeded in constructing a matrix $X$, in an orthonormal
basis, having been initially given a set of distances.

In practice, we might be given dissimilarities rather than distances.
Then, matrix $A$ will be symmetric and have zero values on the
diagonal but will not be positive semidefinite.  In this case
negative eigenvalues are obtained.  These are inconvenient but may
often be ignored if the approximate Euclidean representation (given
by the eigenvectors corresponding to positive eigenvalues) is
satisfactory.  Thus the Euclidean content of the data is quantified using 
the non-negative eigenvalues in the spectral decomposition.   Their 
proportion of the sum of absolute values of eigenvalues is a single 
measure of how metric a data set is.  

\subsection{Quantifying How Ultrametric a Data Set Is}

\subsubsection{Ultrametricity Coefficient of Lerman}

The principle adopted in any constructive assessment of
ultrametricity is to construct an ultrametric on data and see
what discrepancy there is between input data and induced
ultrametric data structure. Quantifying ultrametricity using
a constructive approach is less than perfect as a solution,
given the potential complications arising from known problems,
e.g.\ chaining in single link, and non-uniqueness, or even
inversions, with other methods. The conclusion here is that
the ``measurement tool'' used for quantifying ultrametricity
itself occupies an overly prominent role relative to that
which we seek to measure. For such reasons, we need an
independent way to quantify ultrametricity.

Lerman's \cite{Lerman81} H-classifiability index is as follows.
From the isosceles triangle principle, given a distance $d$ where
$d(x, y) \neq d(y, z)$ we have $d(x, z) \leq 
\mbox{max} \{ d(x, y), d(y, z) \}$,
it follows that the largest and second largest of the numbers
$d(x, y), d(y, z), d(x, z)$ are equal. Lerman's H-classifiability
measure essentially looks at how close these two numbers (largest,
second largest) are. So as to avoid influence of distribution of
the distance values, Lerman's measure is based on ranks (of these
distances) only.  For further discussion of it, see \cite{murtagh04}.

There are two drawbacks with Lerman's index. Firstly, ultrametricity
is associated with H = 0 but non-ultrametricity is not bounded. In
extensive experimentation, we found maximum values for H in the
region of 0.24. The second problem with Lerman's index is that for
floating point coordinate values, especially in high dimensions, the
strict equality necessitated for an equilateral triangle is nearly
impossible to achieve. However our belief is that approximate
equilateral triangles are very likely to arise in important cases
of high-dimensional spaces with data points at hypercube vertex
locations. We would prefer therefore that the quantifying of
ultrametricity should ``gracefully'' take account of triplets
which are ``close to'' equilateral. Note that for some authors,
the equilateral case is considered to be ``trivial'' or a ``trivial 
limit'' \cite{treves}. For us, however, it is an important case,
together with the other important case of ultrametricity (i.e.,
isosceles-with-small-base, which we will write in that way for 
clarity).

\subsubsection{Ultrametricity Coefficient of Rammal, Toulouse and Virasoro}
\label{Rammal}

The quantifying of how ultrametric a data set is by Rammal et al.\ 
\cite{rammal85,rammal86}  
was influential for us in this work. The Rammal ultrametricity
index is given by $\sum_{x,y}(d(x, y) - d_c(x, y))/ \sum_{x,y} d(x, y)$
where $d$ is the metric distance being assessed, and $d_c$ is the
subdominant ultrametric. The latter is also the ultrametric associated
with the single link hierarchical clustering method.  The Rammal et al.\
index is bounded by 0 (= ultrametric) and 1. As pointed out in 
\cite{rammal85,rammal86}  
this index suffers from ``the chaining effect and 
from sensitivity to fluctuations''. The single link hierarchical
clustering method, yielding the subdominant ultrametric, is, as is
well known, subject to such difficulties.

\subsubsection{Ultrametricity Coefficients of Treves and of Hartmann}

Treves \cite{treves} considers triplets of points giving rise to minimal,
median and maximal distances. In the plot of $d_{\mbox{min}}/d_{\mbox{max}}$
against $d_{\mbox{med}}/d_{\mbox{max}}$, the triangular inequality, the
ultrametric inequality, and the ``trivial limit'' of equilateral triangles,
occupy definable regions.

Hartmann \cite{hartmann} considers $d_{\mbox{max}} - d_{\mbox{med}}$. Now,
Lerman \cite{Lerman81} uses ranks in order to give (translation, scale, etc.)
invariance to the sensitivity (i.e., instability, lack of robustness)
of distances. Hartmann instead fixes the remaining distance $d_{\mbox{min}}$.

We seek to avoid, as far as possible, lack of invariance due to use of
distances. We seek to quantify both isosceles-with-small-base
configurations, as well as equilateral configurations. Finally, we seek
a measure of ultrametricity bounded by 0 and 1.

\subsubsection{Bayesian Network Modeling}

Latent ultrametric distances were estimated by Schweinberger and Snijders
\cite{schweinberger} using a Bayesian and maximum likelihood approach
in order to represent transitive structures among pairwise
relationships.  As they state, ``The observed network is generated 
by hierarchically nested latent transitive structures, expressed by 
ultrametrics''.   Multiple, nested transitive structures are at issue.
``Ultrametric structures imply transitive structures'' and
as an informal way to characterize ultrametric structures (arising from
embedded clusters, comprising ``friends'' and  ``close friends''):
``Friends are likely to agree, and unlikely to disagree; close 
friends are very likely to agree, and very unlikely to disagree.''

Issues however in the statistical model-based approach to determining
ultrametricity include that convergence to an optimal fit is not
guaranteed and there can be an appreciable computational requirement.
Our approach (to be described in the next subsection) in contrast
is fast and can be achieved through sampling which supposes that
there is a homogeneous ultrametricity pertaining to the data used.
If sampling is used (for computational reasons) then we assume that
the text is ``textured'' in the same way throughout, or that it is
sufficiently ``unified''.  For one theme in regard to content,
or one origin, or one author, such an assumption seems a reasonable one.

\subsubsection{Our $\alpha_\epsilon$-Ultrametricity Coefficient}
\label{alphae}

We define a coefficient of ultrametricity termed $\alpha_\epsilon$,
parametrized by an approximation factor, $\epsilon$,  which is
specified algorithmically as follows.

\begin{enumerate}

\item All triplets of points are considered, with a distance (by
default, Euclidean) defined on these points. Since for a large number
of points, $n$, the number of triplets, $n(n - 1)(n - 2)/6$ would be
computationally prohibitive, we may wish to randomly (uniformly) sample
coordinates $(i \sim \{ 1..n \}, j \sim \{1..n\}, k \sim \{1..n\})$.

\item We check for possible alignments (implying degenerate triangles) and
exclude such cases.

\item Next we select the smallest angle as less than or equal to 60
degrees. (We use the well-known definition of the cosine of the angle
facing side of length $x$ as: $(y^2 + z^2 - x^2)/2yz.)$ This is our
first necessary property for being a strictly isosceles ($< 60$ degrees)
or equilateral ($= 60$ degrees) ultrametric triangle.

\item For the two other angles subtended at the triangle base, we seek
an angular difference of strictly less than $\epsilon$, defined in our 
past work as 2 degrees or 0.03490656
radians. This condition is an approximation to the ultrametric
configuration, based on an arbitrary choice of small angle. This
condition is targeting a configuration that may not be exactly
ultrametric but nonetheless is very close to ultrametric.

\item Among all triplets (1) satisfying our exact properties (2, 3) and
close approximation property (4), we define our ultrametricity coefficient
as the relative proportion of these triplets. Approximately ultrametric
data will yield a value of 1. On the other hand, data that is
non-ultrametric in the sense of not respecting conditions 3 and 4 will
yield a low value, potentially reaching 0.
\end{enumerate}

In summary, the $\alpha_\epsilon$ index is defined in this way:

Consider a triplet of points, that defines a triangle. If the
smallest internal angle, $a$, in this triangle is $\leq 60$ degrees,
and, for the two other internal angles, $b$ and $c$, if $ |b - c| < \epsilon$
radians, then this triangle is an ultrametric one. We look for the
overall proportion of such ultrametric triangles in our data.

\subsection{Application to Partial Ultrametric Embedding}

In \cite{steklov}, we discuss
permutation representations of a data stream.  Since hierarchies
can also be represented as permutations, there is a ready way to
associate data streams with hierarchies.  In fact, early computational
work on hierarchical clustering used permutation representation to
great effect (cf.\ \cite{sibson}).

In \cite{murtaghEPJ}, in analysis of time series, chronological or other 
ordered 1-dimensional signals, data was fingerprinted in the domains of 
chaotic (found to be less ultrametric), biomedical, meteorological, financial
data.
To analyze data streams in this way, in
\cite{murtaghEPJ} we developed an approach to ultrametric embedding of
time-varying signals, including biomedical, meteorological, financial
and other.  As opposed to the classical way of inducing a hierarchy,
through use of an agglomerative hierarchical clustering algorithm,
we looked for the ultrametric relationship -- the strong triangular
inequality -- and, when found, counted such particular cases of adherence
to inherent hierarchical properties in the data.  The most non-ultrametric
time series are found to be chaotic ones.   Eyegaze trace data was found
to be remarkably high in ultrametricity, which is likely to be due to
extreme saccade movements.  Some initial questions were raised in
that work (\cite{murtaghEPJ}) in regard
to the EEG data used, for sleeping, petit mal and irregular epilepsy cases.

This work was pursued by Khrennikov and his colleagues in
modelling multi-agent systems.   See \cite{khrennikov2008}.  Furthermore
this work used Bose-Einstein and Fermi-Dirac statistical distributions
(derived from quantum statistics of energy states of bosons and fermions, i.e.\
elementary particles with integer, and half odd integer, spin).
In \cite{khrennikov2006,khrennikov2008} multi-agent
behaviours are modelled using such energy distributions.  The framework
was an urn model, where balls can move, with loss of energy over time,
and with possibilities to receive input energy, but potentially shared with other
balls.  See the cited works for a full description of the Monte Carlo system
set up.  Sequences of actions (and moves), viz.\ their histories,
 are coded such that triangle
properties can be investigated (cf.\ also \cite{murtaghEPJ}).  That leads
to a characterization of how ultrametrically-embeddable the data is, {\em ab              
initio} (and not through imposing any hierarchical or other structure on the
data with retrospective goodness of fit assessment).
In \cite{khrennikov2008}, the case is presented for such analysis of
behavioural
histories being important for study of social and economic complexity.

Quantum statistical distributions have been noted in the foregoing
work (\cite{khrennikov2006,khrennikov2008}).  Van Rijsbergen
\cite{vanrijsbergen2004} has set out various ways in which a quantum
physics formalism makes clearer what is being done in information
retrieval and in data analysis generally.

The quantifying of the inherent ultrametric content of text, and finding
that some are much more inherently hierarchical than others, was pursued
in \cite{murtaghmind2}.  In this way, a large number of texts were 
fingerprinted.  As data, the following were used: tales from the
Brothers Grimm, Jane Austen novels, dream reports, air accident reports,
and James Joyce's Ulysses.

\section{Transforming Data to Become More Metric or More Ultrametric}

\subsection{Transforming Non-Metric to Metric Data}

We recall that a metric satisfies symmetry and positive definiteness,
together with the triangular inequality: $\forall i, j, k, d(i,j) = 
d(j,i); d(i,j) \geq 0, d(i,j) = 0 \Longleftrightarrow i = j; d(i,j) 
\leq d(i,k) + d(k,j)$.  A dissimilarity does not satisfy the triangular
inequality.  Hence a dissimilarity is not directly embedable in a metric
space, such as a Euclidean 2D (planar) or 3D space typically used for
visualization.  Next we consider (i) mapping metric data to a best low
dimensional projection, (ii) mapping non-metric, hence dissimilarity,
data to a best low dimensional metric space, and (iii) transforming
non-metric data into close if not best-fitting metric values.

Consider a set $I$ endowed with a dissimilarity, $d_I$.  Multidimensional
scaling is an approach to take $d_I$ into a distance, $\delta_I$, which
therefore as a distance respects the triangular inequality.  In
multidimensional scaling it is just the ranks of $d_I$ that are used
(so $d_{ij} < d_{kl}$ for all couples $(i,j)$ and $(k,l)$.

In Torgerson's \cite{torgerson} metric multidimensional scaling, a low dimensional
representation of Euclidean data is obtained.  The least-squares best
fit is provided by the orthonormal axes that define the principal axes
of inertia.  In non-metric multidimensional scaling \cite{kruskal},
dissimilarities of a general sort, that are rank ordered, are
mapped non-linearly into a low dimensional Euclidean representation.
A goodness of fit measure between input dissimilarities and output
distances in the low dimensional space, called ``stress'', is used
as an optimand.  Iterative, steepest descent is used for optimization.
Additional short description is to be found in \cite{murtaghheck,
murtagh2000}.

There are other ways of taking a dissimilarity into a distance.
Consider the following from Cailliez and Pag\`es (\cite{cailliez} p.\ 257).  The
most common are from the following family, where $r$ is a positive value:

$$\delta_{ij} = 0 \mbox{ if } i = j \ \mbox{ ; } \ 
 d_{ij} = (d^r_{ij} + c)^{1/r} \mbox{ if } i \neq j$$
where $c$ is a positive constant chosen so that the triangular inequality
is respected by the function $\delta_I$.

For $r = 1$, taking:

$$c = \max \{ d_{ij} - d_{ik} - d_{jk} | i, j, k \in I^3 \} $$
which is the minimum possible value for $c$.  This provides for a
simple transformation approach.

Many other transformations are possible, such as for example:

$$\delta_{ij} = d_{ij}^r \ \ \mbox{  for all  } (i,j) \in I \times I  $$
where $r$ is the largest value between 0 and 1 such that $\delta_I$
satisfies the triangular inequality.

\subsection{Transforming Metric, or Non-Metric, to Minimal Superior
Ultrametric, and to Maximal Inferior Ultrametric}

In section \ref{AHC}, agglomerative hierarchical clustering was 
described.  

Rather than the triangular inequality, ultrametric data are characterized
by the properties of a metric together with the ultrametric, or 
strong triangular, inequality,
$d_u(i,j) \leq \mbox{max} \{ d_u(i,k), d_u(j,k) \}$.  Hierarchical agglomerative
clustering algorithms are a general and widely-used class of algorithm
for inducing an ultrametric on dissimilarity or distance input, or
coordinate data on which a metric or a dissimilarity has been defined.

Johnson \cite{john} makes every triangle isosceles by making its smallest
side the base.  This leads to the maximal inferior ultrametric, also known
as the subdominant ultrametric, or ultrametric produced by the single
linkage hierarchical clustering agglomerative algorithm.  See
Benz\'ecri \cite{benz} (p.\ 144).
The maximal inferior ultrametric is unique.

The minimal superior ultrametric is not though.  
In this case, Benz\'ecri \cite{benz} 
(p.\ 144) 
discusses the adaptive modification of triangle edges to enforce
the ultrametric inequality, i.e.\ isosceles triangles with small base.   

\subsection{Approximating an Ultrametric for Similarity Metric Space Searching}
\label{stretch}

In \cite{murtagh04} we show that, in
much work over the years, nearest neighbour searching has been made
more efficient through the use of more easily determined feasibility bounds.
An early example is Fukunaga and Narendra \cite{fukunaga}, a chapter review 
is in
\cite{Murtagh85-1}, and a survey is to be found in \cite{Chavez01}.
Rendering given distances as
ultrametric is a powerful way to facilitate nearest neighbour searching. 
Furthermore
``stretching the triangular inequality'' (a phrase used by \cite{chavez03})
so that
it becomes the strong triangular inequality, or ultrametric inequality, gives a
unifying view of some algorithms of this type.

Fast nearest neighbour finding often makes use of pivots to establish bounds
on points to be searched, and points to be bypassed as unfeasible
\cite{Bustos03,Chavez01}.

A full discussion can be found in \cite{murtagh04}.
Fast nearest neighbour searching in metric spaces often appeals to heuristics.
The link with ultrametric spaces
gives rise instead to a unifying view.

Hjaltason and Samet \cite{hjaltason} discuss
heuristic nearest neighbour searching in terms of embedding the given metric
space points in lower dimensional spaces. From our discussion in this section,
we see that there is evidently another alternative direction for 
facilitating fast nearest neighbour searching: viz., taking the metric 
space as an ultrametric one,
and if it does not quite fit this perspective then ``stretch'' it 
(transform it locally) so that it does so.

\section{Emotion as the Doorway into the Subconscious}

\subsection{Emotion as an Essential Component of Perception}

In \cite{murtaghmind1,murtaghmind2} we have described how Matte Blanco's
psycho-analytical theories lend themselves very well to the following 
viewpoint, a viewpoint that was noted first in \cite{laurogrotto}.  
If a Euclidean, or more generally metric, space is a good framework 
for time, causality, semantics (e.g.\ through all pairwise distance 
relationships, cf.\ \cite{murboole}), and Matte Blanco's {\em asymmetric} 
thought processes (see e.g.\ \cite{murtaghmind1}), then metric space
collaterally is useful for mapping conscious activities.  By implication,
metric space is a good model for consciousness.  On the other hand,
ultrametric spaces are a good way of envisioning Matte Blanco's 
{\em symmetric} thought 
processes, atemporal thoughts, and episodic (as opposed to semantic 
memory, for example in studies of Alzheimer's disease, see 
\cite{marcotonti}).  

In the PhD thesis of Marco Tonti, \cite{marcotonti}, 
Matte Blanco's work is taken in 
the domain of human emotion, which is closely associated with the 
human subconscious.  Emotion is ``the component of thought that can be
attributed to [...] unconscious rules'' (\cite{marcotonti}, p.\ 78). 
His aim in his thesis is to demarcate that component of thought.  
Since tangible objects are imbued with emotion (``an object in a bag of 
symmetry which is deeply irradiated with emotion'', and we can go a lot
further in considering ``a highly emotionated object'', pp.\ 87, 88), 
the problem addressed by Tonti is that 
``...some mechanism of connecting the experiential/situational dimension
of an individual with the deep/unconscious dimension, in terms of
emotionated objects, is required.'' (p.\ 87).  

Furthermore, 
``The fundamental theoretical hypothesis of this
dissertation is that emotions are an essential component in the process of
perception (and therefore evaluation) of the world. Consequently, the
amount of emotions involved in such a task is expected to be reflected in
the way of categorizing/expressing evaluations of a subject.''  (p.\ 119).

An important point made is that 
``...emotions are a fundamental and
building force for cognition, and not just an attribution of affective
aspects or a physical response.'' (p. 164).
Emotions for Tonti are a fundamental constituent of thought and 
intelligence.  They are not juxtaposed but rather are an essential 
part of the latter domains, viz.\ thought and intelligence.   The common 
perception is to take 
emotions as ``the opposite'' (in some sense) of thought.  However, 
for Tonti, intelligence would not be possible without emotion.  
Emotions are not just the product of an appraisal by us of some thing, 
but the very reason for which some elements of the world are relevant. 
Emotions are also regulatory functions in our (human) process of
segmenting reality.  (p.\ 9).  An emotion is not just a datum
but is constitutive of our internal reality.   

In his objective of having ``emotion measuring techniques'', Tonti notes 
that emotion is usually taken as a {\em reaction} to events, and this can
be assessed in some physical manner, such as via a report by the subject, 
biomarkers and bodily signals, etc.  

Counterposed to this reflection of emotion, Tonti sets out a more ambitious
aim (p.\ 122): 
``In the rest of the chapter are presented two ways of measuring emotions
conceptually based on the unconscious functioning of the mind, which
seeks for the emotional activity not in the presumed emotional responses
to an arbitrary stimulus, but in the influence of emotionality in the way of
responding. If emotions are a component of thought, then it would be
possible to observe the trails of their presence and influence, in the
diverse degrees, in the overall process of mind functioning.''  

The big advantage of posing the problem of demarcating emotion in this 
way is that it opens up better vantage points on tracking emotion, and
on emotion in context (of other emotions, or emotionated objects, of the 
subject's mind processes, and so on).   

Furthermore Tonti considers metric, as well as ultrametric, 
mathematical frameworks.   

``The metric content index is therefore a measure of the amount
of structure embedded in the neural representations that
inform subject choice: It is high when individual memory items
are classified using semantic cues, which leads to a more
concentrated distribution of errors. It is low either when
performance is random (in which case performance measures
are also low), or when episodic access to the identity of each
famous face is prevalent, semantic relationships remain largely
unused, and errors, when made, tend to be more randomly
distributed.'' (p.146).  The foregoing is with reference to the 
balance or strength of semantic memory versus episodic memory, useful 
in observing Alzheimer's patients (where the latter increases in 
dominance; see \cite{marcotonti}, p.\ 14 and elsewhere).   

Then the ultrametric understanding of the subconscious is 
described in these terms: 
``...an important idea is the one that
sees the unconscious in terms of a topological semantic structure with
specific features. If we imagine the experience of a life as encoded in a
network of representations of facts, ideas, relations and so on, it can
hypothesized that some specific distance between the elements can be
defined. In this novel interpretation [...] brought forward by
Lauro-Grotto (called ``ultrametric''), the fabric of this network is modified
in a way that, in certain circumstances, the distance between a group G of
otherwise distinct objects and a third object X is considered to be the
same for each of them. If we consider the distance as the probability of
going from X to each of the object in G, we should conclude that the
probability of going from X to any of the objects in G is the same, i.e.\ they
are structurally considered to be equivalent.''   (p. 129)

This work of \cite{marcotonti} is motivation for the new work that follows.  
We will seek to directly point out emotionated objects, and rank them in 
accordance with how well they present a characteristic related to the 
subconscious.  This characteristic is that of being inherently 
ultrametric.

\subsection{Data With Relatively High Ultrametricity}

From the Dreambank repository 
\cite{domhoff2002,domhoff2003,dreambank2004,schneider}
we selected various collections.  See Murtagh \cite{murtaghmind1,murtaghmind2}
for 
further description and analyses.   One set of 139 dream reports, from one 
individual, Barbara Sanders, was particularly reliable (according to 
\cite{dreambank2004}). 

In order to have a text that ought to contain vestiges of ultrametricity because of subconscious
thinking, admittedly subconscious thinking that was afterwards reported on in a fully conscious way,
we took these Barbara Sanders dream reports. 
In discussion of this data provided in Domhoff 
\cite{domhoff2002} he notes that there is ``astonishing
consistency'' shown in dreams such as these over long periods of time.
Taking the set of 139 of the Barbara Sanders dream reports, we used the 2000
most frequently occurring words used in these dream reports 
including function words. Then we took
30 words to carry out some experimentation with their ultrametric 
properties. These are listed as follows.  

Note that our processing converts all upper case to lower case.
The thirty words selected were as follows: 

``tyler'', ``jared'', ``car'', ``road'', ``derek'', ``john'', ``jamie'',  
``peter'', ``arrow'', ``dragon'', ``football'', ``lance'', ``room'', ``bedroom'',
``family'', ``game'', ``mabel'', ``crew'', ``director'', ``assistant'', ``balloon'',
``ship'', ``balloons'', ``pudgy'', ``valerie'', ``dolly'', ``cat'', ``gun'', ``howard'',
``horse''

We selected these words to have some personal names, some words that could be metaphors for the
commonplace or the fearful, and some words that could be commonplace and hence banal.
Names of people are: Tyler, Jared, Derek, John, Jamie, Peter, Lance, Mabel, Valerie,
and Howard. 

\begin{figure}
\begin{center}
\includegraphics[width=14cm]{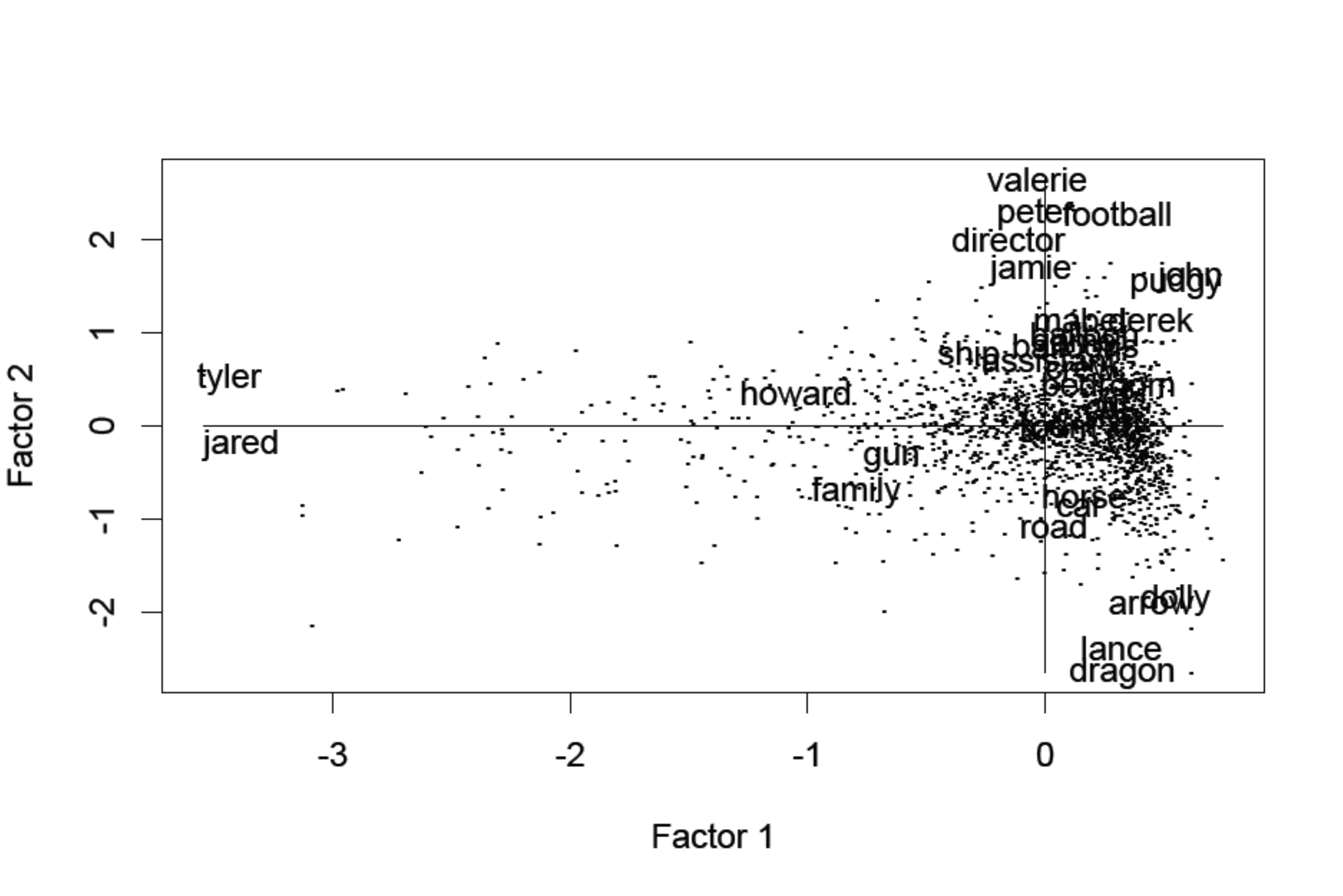}
\end{center}
\caption{2000 terms, 30 of them indicated - see how tyler and jared; lance 
and dragon; valerie, football, peter, director, jamie; etc. come out close.
Note this is a planar approximation -- 2.2\% and 1.37\% of the inertia 
explained by these factors, respectively.}
\label{figca}
\end{figure}

We carried out, firstly, a Correspondence Analysis of all 139 dream reports crossed by 
the 2000-word set.  Then we use for the subsequent analysis the Euclidean, factor space,
with full dimensionality, for the 30 selected terms.  Figure \ref{figca}
shows the principal factor plane projection.  

The full Euclidean, factor, space dimensionality is the minimum of 
$2000 - 1$ and $139 - 1$, hence: 138.  By selecting 30 
terms of interest, our data to be further explored is a table that crosses
30 chosen terms by 138 Euclidean space coordinates, viz.\ the factor 
projections.  The projections on factors 1 and 2 are to be seen in Figure 
\ref{figca}.  

\subsection{A New Principle for Metric and Ultrametric Component Analysis}

In seeking ultrametric relationships in data, our approach in 
section \ref{quantifying} has been to single out triplets of the 
observables that respect the necessary ultrametric properties. 

In section \ref{transforming}, the well established data analytic
approach imposes an ultrametric topology on data, and then 
interprets this retrospectively.

The former, quantifying, approach does not give us insight into what 
could become part of the ultrametric component in the data, were it 
possible to bend (or ``stretch'', the term used in subsection 
\ref{stretch}) the data in order for the data to become more ultrametric.

The latter approach suffers from having any number of different 
hierarchies that can be induced on the data, in accordance with 
criteria (or definitions) for doing so.   

We will introduce a new approach to determining the ultrametric 
component of a data set, by combining both quantifying and transforming
approaches.  We do so as follows. 

\begin{enumerate}
\item Determine an ultrametric consensus space of given ultrametric 
spaces.  Operationally, we fit hierarchies, using different agglomerative
criteria, to out data, and we determine a consensus hierarchy.
\item From the consensus hierarchy, implying a mapping of our data
into this ultrametric topology, we determine with respect to our original,
input measurements, just how close the mapped ultrametric relationships
are to our input relationships.  We retain the better mapped relationships.
\end{enumerate}

In sections \ref{consensus} and \ref{seceps} we discuss these two 
stages of our new principle of ultrametric component analysis.  

\section{Consensus Ultrametric Sets from Two Hierarchies}
\label{consensus}

\subsection{Ultrametric Consensus: Definition and Algorithm}

In Appendix A1, there is a short review of consensus approaches for 
hierarchical clustering.  Here we pose the following consensus problem.
Firstly we note how we define ultrametric relations on the basis of any 
given triangle of observations or points.  Therefore, in seeking a 
consensus between hierarchies, we start by checking each triangle defined
from the terminals, that are associated with our observations.  

In an agglomerative hierarchical clustering with pairwise agglomerations,
and assuming that no input distances are identical (which will not 
always be respected in practice; however our assumption of unique
input distances is aided by the Euclidean, hence continuous, values
taken to start with), then it follows that ultrametric cases associated
with equilateral triangles will not arise.  
The isosceles-with-small-base case 
is the only one that is applicable.  Hence, in our 
considerations here we exclude tied
distances that would lead to no smallest base side: we require that there
be a ``base'' side in any triplet.  

Our ultrametric consensus algorithm is as follows:

\begin{enumerate}
\item Consider all triplets of observations (i.e., singletons associated
with terminal nodes in the dendrogram), $i, j, k$. 
\item Consider also the cophenetic matrix, i.e.\ the matrix of ultrametric
distances, $d_u$.  Read off: $d_u(i,j), d_u(j,k), d_u(i,k)$.  Determine:
$d_u^{(1)} < d_u^{(2)} = d_u^{(3)}$.  
\item Next consider that two dendrograms produced by agglomerative 
hierarchical cluster are at issue.  Determine for each:

$d_{u1}^{(1)} < d_{u1}^{(2)} = d_{u1}^{(3)}$

$d_{u2}^{(1)} < d_{u2}^{(2)} = d_{u2}^{(3)}$

\item Recapitulate: we have selected the same triplet, $i, j, k$, in the 
two hierarchies.  We look at the ultrametric relationship between the 
three points, $i, j, k$.  These three points must have the property of 
forming an isosceles triangle with small base.  Without loss of generality 
we ignore the case of an equilateral triangle.  (Why is this justified? 
Firstly, due to each agglomeration in the hierarchical agglomerative 
algorithm taking two nodes at a time; secondly, given the real values 
of distance or dissimilarity.)  

Let $i$ be the triangle apex in the first
hierarchy, and $j$ and $k$ then are the points at the small triangle base.
In that case, the distance between $j$ and $k$ is the small base side of 
the triangle connecting the three points.  We need to see if the same 
triangle apex point is present in the second hierarchy.  That is all we 
need to do, in order to have a morphological equivalence between the two 
cases of ultrametricity, extracted from the hierarchies. 

\item For the ultrametric relationships associated with the two 
hierarchical clusterings to be the same, we require that the base 
side of the triangles involve the same observations in the two 
cases: 

Given the lowest ranked triangle side in the two cases: 

$d_{u1}^{(1)}(i_1,i_2), d_{u2}^{(1)}(i_3,i_4)$

we require that the base side involve the same observations: 
either $i_1 = i_3; i_2 = i_4$ or $i_1 = i_4; i_2 = i_3$.


\item Count these matches between the hierarchies.  
The count is that of isosceles-with-small-base that are consistent for the 
two hierarchies.  The total number of triangles considered, for $n$ 
observations (hence $n$ terminal nodes in the hierarchical tree), is 
$n \cdot (n - 1) \cdot (n - 2) / (3 \cdot 2 \cdot 1)$.  This furnishes
a coefficient of ultrametric consensus between the two hierarchies, 
or ultrametric embedding of the same set of observations.

\item For all isosceles-with-small-base cases, these determine a consensus
between the two hierarchies.  The consensus is the set of triplets of points
(i.e.\ singleton clusters, or terminals in the dendrogram, or observations)  
that (i) respect the ultrametric condition of having a triangle with 
small base -- which is a necessity, given the hierarchical or tree 
structure of the data; and (ii) that have the same apex point in the two 
hierarchies considered.   

\end{enumerate}  

\subsection{Application: Ultrametric Consensus of Hierarchies}

\begin{figure}
\begin{center}
\includegraphics[width=14cm]{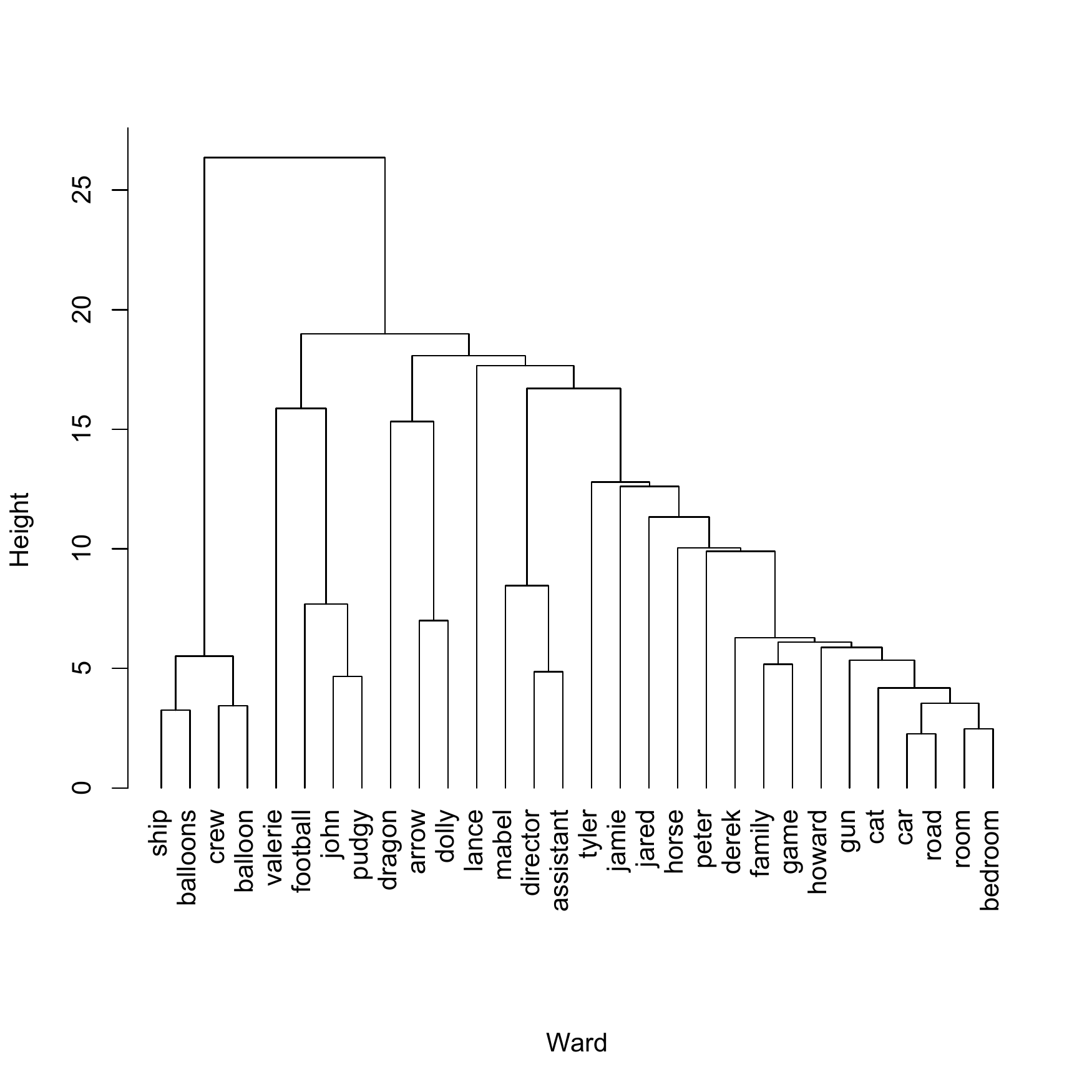}
\end{center}
\caption{Hierarchical clustering using the Ward minimum variance 
agglomerative criterion.  (Relative to Figures \ref{fig3hc2}, \ref{fig3hc3}, 
and
just as one example,  the similar clustering sequence
of ``ship'', ``balloons'', ``crew'', ``balloon'' is to be noted in all
three cases.)}
\label{fig3hc1}
\end{figure}

\begin{figure}
\begin{center}
\includegraphics[width=14cm]{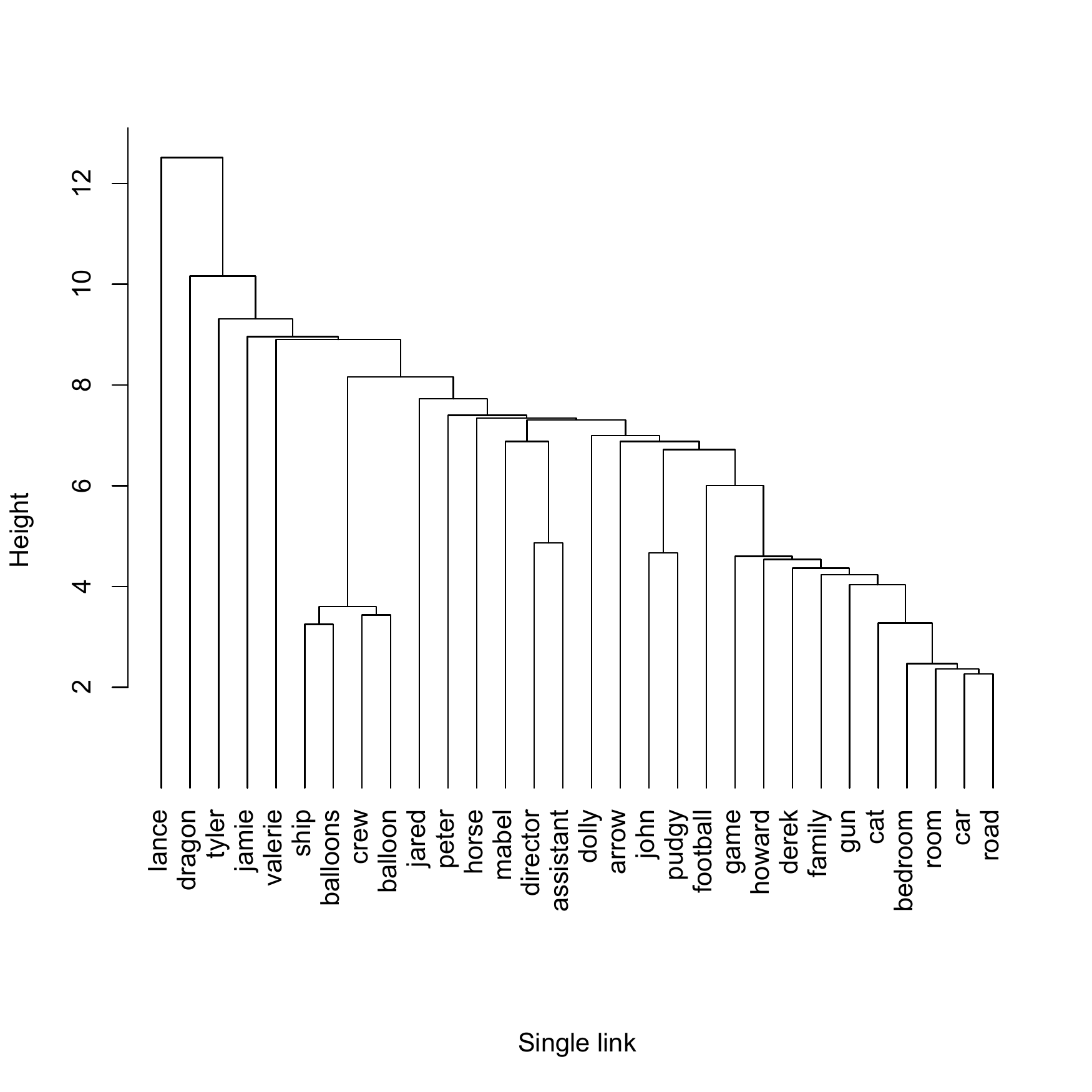}
\end{center}
\caption{Hierarchical clustering using the single link  
agglomerative criterion.  (Relative to Figures \ref{fig3hc1}, \ref{fig3hc3}, 
and just as one example,  the similar clustering sequence
of ``ship'', ``balloons'', ``crew'', ``balloon'' is to be noted in all
three cases.)}
\label{fig3hc2}
\end{figure}

\begin{figure}
\begin{center}
\includegraphics[width=14cm]{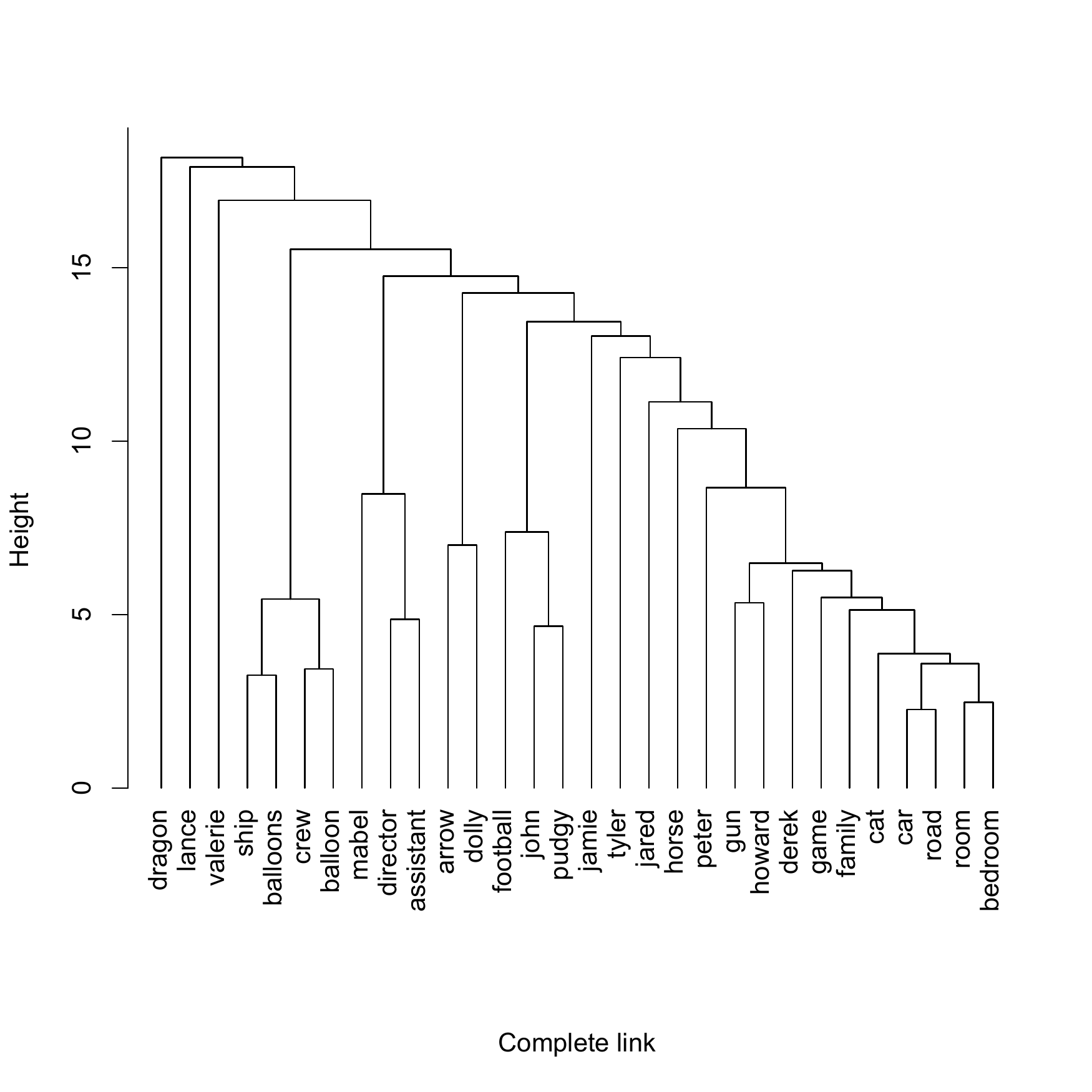}
\end{center}
\caption{Hierarchical clustering using the complete link  
agglomerative criterion.  (Relative to Figures \ref{fig3hc1}, \ref{fig3hc2}, 
and
just as one example,  the similar clustering sequence
of ``ship'', ``balloons'', ``crew'', ``balloon'' is to be noted in all
three cases.)}
\label{fig3hc3}
\end{figure}

\begin{figure}
\begin{center}
\includegraphics[width=14cm]{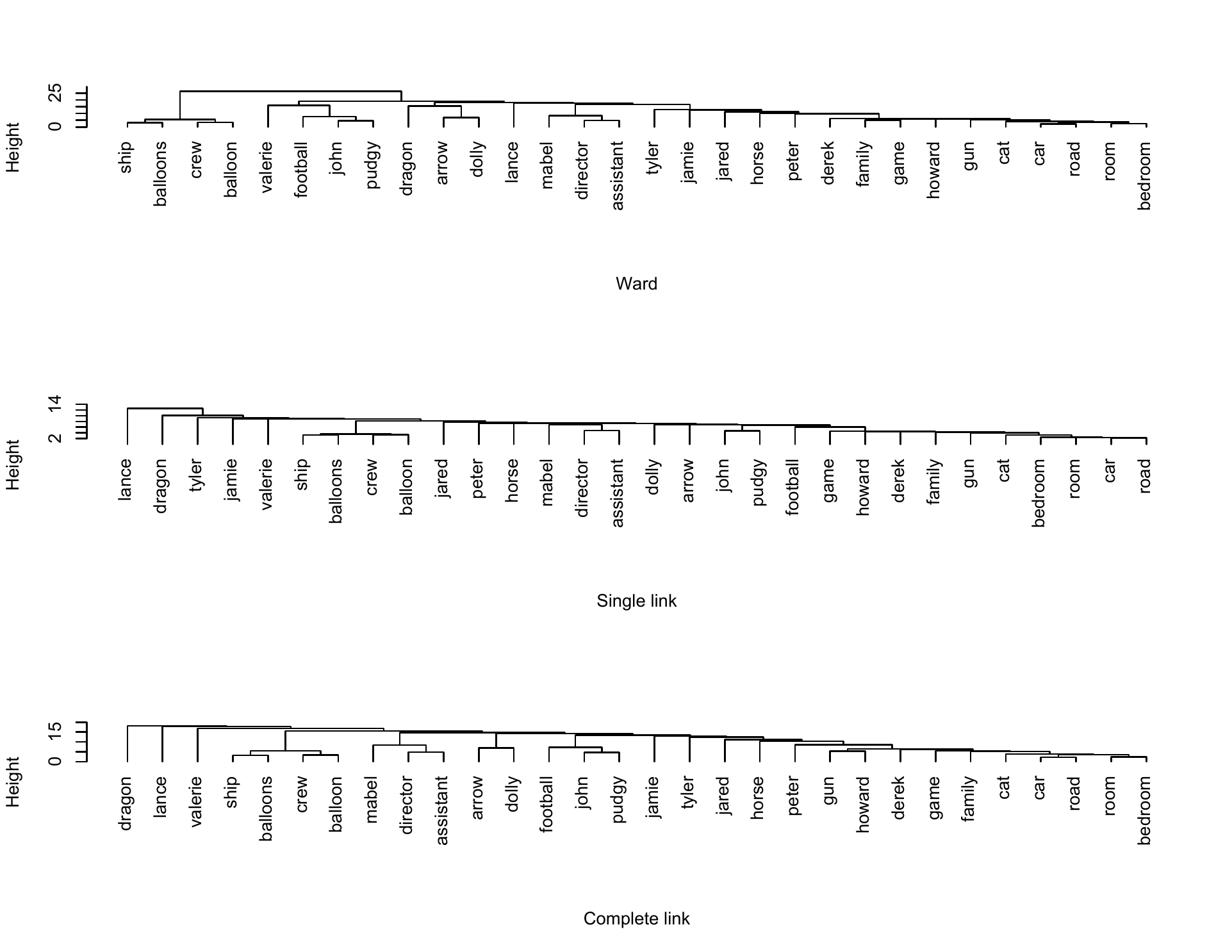}
\end{center}
\caption{For ease of comparison, this shows together
 Figures \ref{fig3hc1}, \ref{fig3hc2}, \ref{fig3hc3}.  
Hierarchical clusterings using (top) the Ward minimum variance 
agglomerative criterion, (middle) the single link method, and (bottom)
the complete link method.  (Just as one example,  the clustering
of ``ship'', ``balloons'', ``crew'', ``balloon'' is to be noted in all
three cases.)}
\label{fig3hc}
\end{figure}

Figure \ref{fig3hc} shows hierarchical clusterings with three agglomerative 
criteria.  The ultrametric relationship consistency between these hierarchies
was found to be the following.  Firstly, there are 4060 triangles in total.
For Ward and Single link, there were 2369 matching isosceles-with-small-base
triangles.  For Ward and Complete link, there were 3084 matching ultrametric
cases.  For Single link and Complete link, there were 3128 matching cases. 

\subsection{Inversions in Hierarchical Clustering}

A requirement for our ultrametric consensus algorithm is that 
the hierarchy does not contain inversions.  In section \ref{consensus}
in step 4 it is necessary that the lowest ranked triplet (or triangle)
side is the distance associated with the first (in the sequence of
agglomerations) agglomeration between observations.  When there 
are inversions in the hierarchy, which we will now exemplify,
this first agglomeration involving triplet vertices may be associated
with a distance, hence triangle side, that is not the lowest ranked.

\begin{figure}
\begin{center}
\includegraphics[width=12cm]{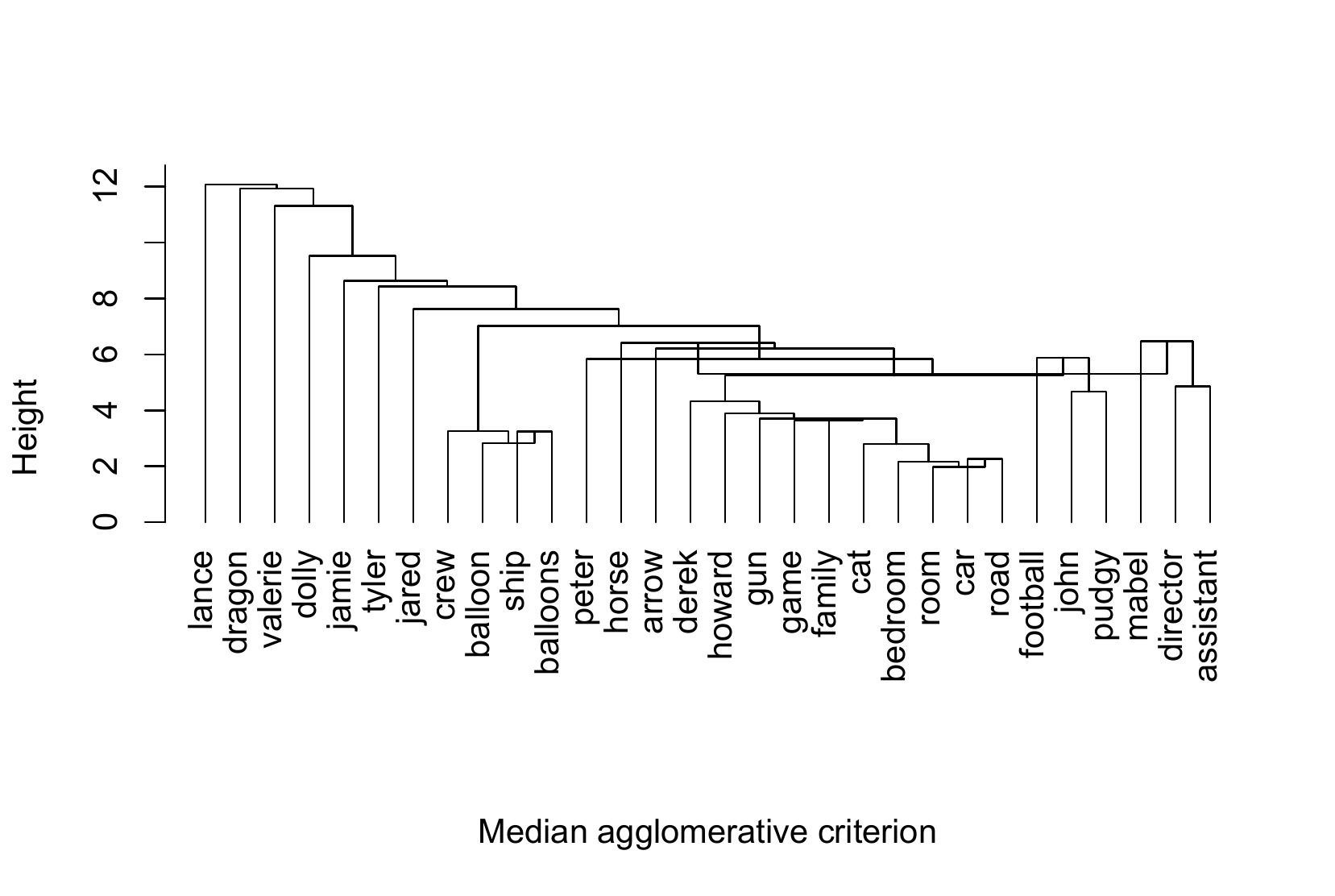}
\end{center}
\caption{Inversions in the sequence of agglomerations.  That is, 
$i$ and $j$ merge, and the distance of the this new cluster to another 
cluster is {\em smaller} than the defining distance of the $i, j$ 
merger.  Hence, there is non-monotonic change in the level index, given 
by the distance defining the merger agglomeration.}
\label{figinv}
\end{figure}

Figure \ref{figinv} shows quite a few inversions in the sequence
of agglomerations, meaning that there is non-monotonic increase in 
inter-cluster (cluster encompassing singleton cluster also) distance
as the agglomerations are carried out.  Inversions occur for 
particular agglomerative criteria.  See \cite{Murtagh85-1}.  These 
include the centroid and median criteria.  These therefore must
be avoided in such ultrametric consensus finding since the 
base of the isosceles triangle is not necessarily found with 
such hierarchies.  

A condition for a hierarchy to be guaranteed not to give rise to 
inversions was provided in \cite{bruynooghe}.  This is formulated 
in one form of Bruynooghe's {\em reducibility property} 
\cite{bruynooghe} as:

$$   \mbox{ Inversion impossible if: }                         
      d(i,j) < d(i,k) \  {\rm or} \   d(j,k)   \Rightarrow 
      d(i,j) < d(i \cup j,k)$$

\subsection{Ultrametric Consensus}

Taking the 30 selected terms, originally in a 2000-dimensional 
space that leads to the semantic space that jointly represents 
the projected terms and observations (the latter: dream reports),
we find the following for the number of consistent ultrametric
triplets from the total set of 4060 triplets.

To recapitulate the ultrametric consensus algorithm used for the
two hierarchies, using their derived cophenetic, i.e.\ ultrametric,
distances: 

For each triplet, take the pairwise ultrametric (cophenetic) distances; 
rank them, for the triplet in both cases; the ranks are 1, 2.5, 2.5 for 
the case of isosceles-with-small-base; rarely but possibly the ranks are
2, 2, 2 for the case of an equilateral triangle.   Knowing now the 
small side of the isoceles triangle, check if this is based on the 
same pair (of points in the triplet, hence of triangle vertices) in the
case of the two triplets.  If it is, we have ultrametric consistency between
the two triplets.  

\begin{table}
\begin{tabular}{lrrrrr} \hline
        & Ward & Average & Single & Complete & McQuitty \\ \hline 
Ward    & 4060 & 3050 & 2369 & 3084 & 3014 \\
Average & 3050 & 4060 & 3311 & 3480 & 3830 \\
Single  & 2369 & 3311 & 4060 & 3128 & 3336 \\
Complete & 3084 & 3480 & 3128 & 4060 & 3327 \\
McQuitty & 3014 & 3830 & 3336 & 3327 & 4060 \\ \hline
\end{tabular}
\caption{Hierarchies -- see Figures \ref{fig3hc1}, \ref{fig3hc2},
\ref{fig3hc3} -- are compared by looking at each triplet.  Given
that each hierarchy uses the same set of $n = 30$ observations, there
are 
$n \cdot (n-1) \cdot (n-2) / ( 3 \cdot 2 \cdot 1) = 4060$ 
triplets.
The values in this table give the numbers of isoceles-with-small-base
triplets found, that are morphologically consistent, i.e.\ with the same
triangle base pair, across the two hierarchies compared.}  
\label{tabcons}
\end{table}

Table \ref{tabcons} shows that a great deal of consensus can be 
obtained from the hierarchies.  Comparing a hierarchy against itself,
as done for the main diagonal elements of the table, is just done to 
find what the expected full consistency.  

\begin{table}
\begin{tabular}{lrrrrr} \hline
        & Ward & Average & Single & Complete & McQuitty \\ \hline 
Ward    & 4060 & 2103 & 1714 & 2478 & 1978 \\
Average & 2103 & 4060 & 3608 & 1762 & 2955 \\
Single  & 1714 & 3608 & 4060 & 1500 & 2770 \\
Complete & 2478 & 1762 & 1500 & 4060 & 2134 \\
McQuitty & 1978 & 2955 & 2770 & 2134 & 4060 \\ \hline
\end{tabular}
\caption{Using randomly generated data (see text for details), the table
shows the numbers of morphologically consistent, vertex-labelled,  
ultrametric triangles between pairs of hierarchies.}
\label{tabconsran}
\end{table}

In Table \ref{tabconsran} we benchmark the Table \ref{tabcons} results
by taking some random data.  

We ``mirrored'' the real data used throughout this work in the following
way.  First, we generated a matrix with uniformly distributed values on 
[0,1], of dimensions 139 $\times$ 2000.  We mapped it into a Euclidean 
representation, for full consistency of treatment with the data used
throughout this work.  Then using the 
full dimensionality, 138 (i.e.\ one less than the minimum of the row 
and column dimensions of the input data matrix) 
 of the Euclidean, Correspondence Analysis mapping into
the Euclidean space, we selected 30 vectors.  (How we did this was 
as follows.  We used exactly the sames indices as used in our real 
data, selecting from the 2000 vectors.  Hence, we replicated what was 
done on the real data, in order to furnish a comparable benchmark in 
all respects.)

\begin{figure}
\begin{center}
\includegraphics[width=14cm]{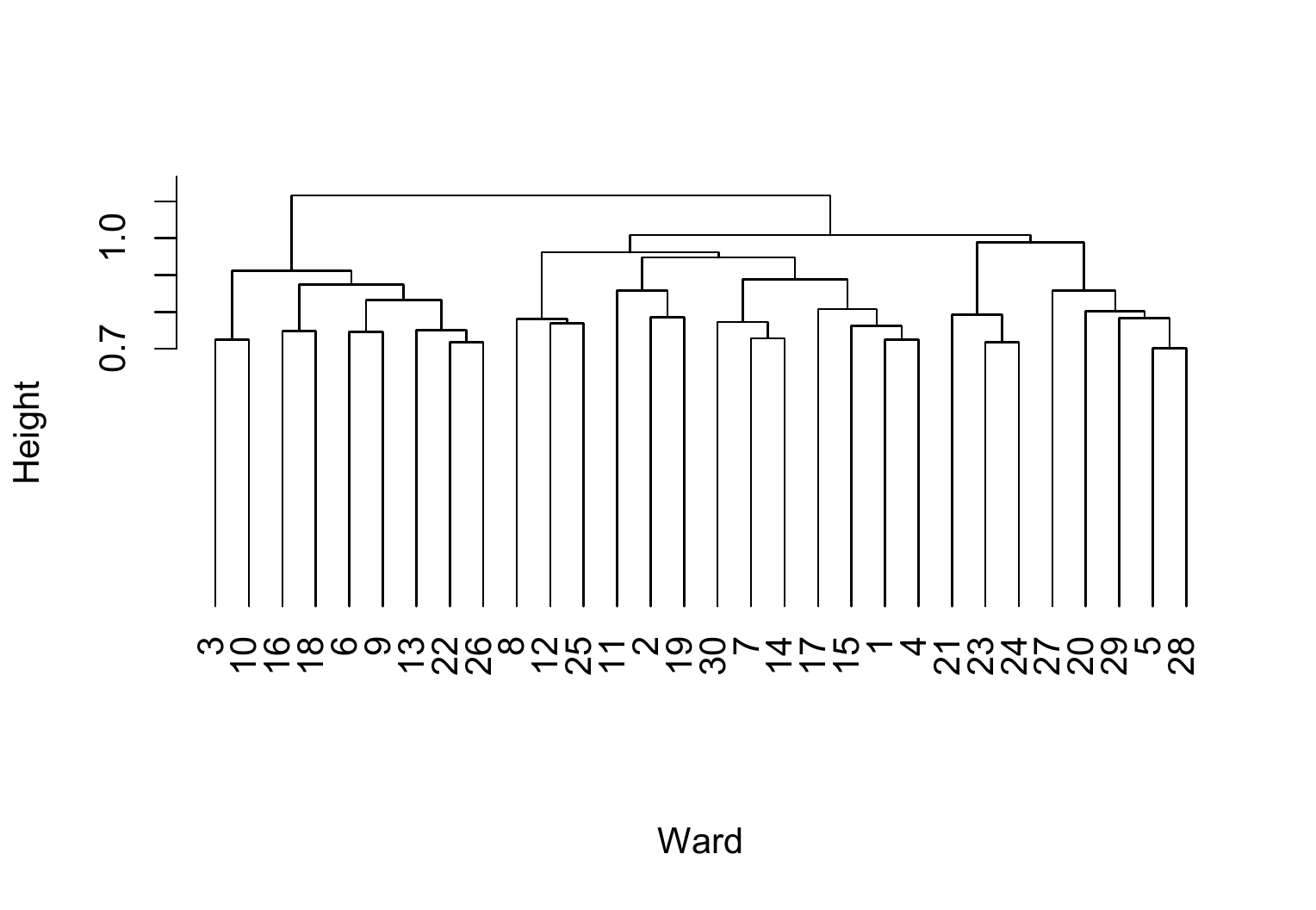}
\end{center}
\caption{Hierarchical clustering using the Ward minimum variance 
agglomerative criterion.  Input data are randomly generated.  
(See text for details.)} 
\label{fighran1}
\end{figure}

\begin{figure}
\begin{center}
\includegraphics[width=14cm]{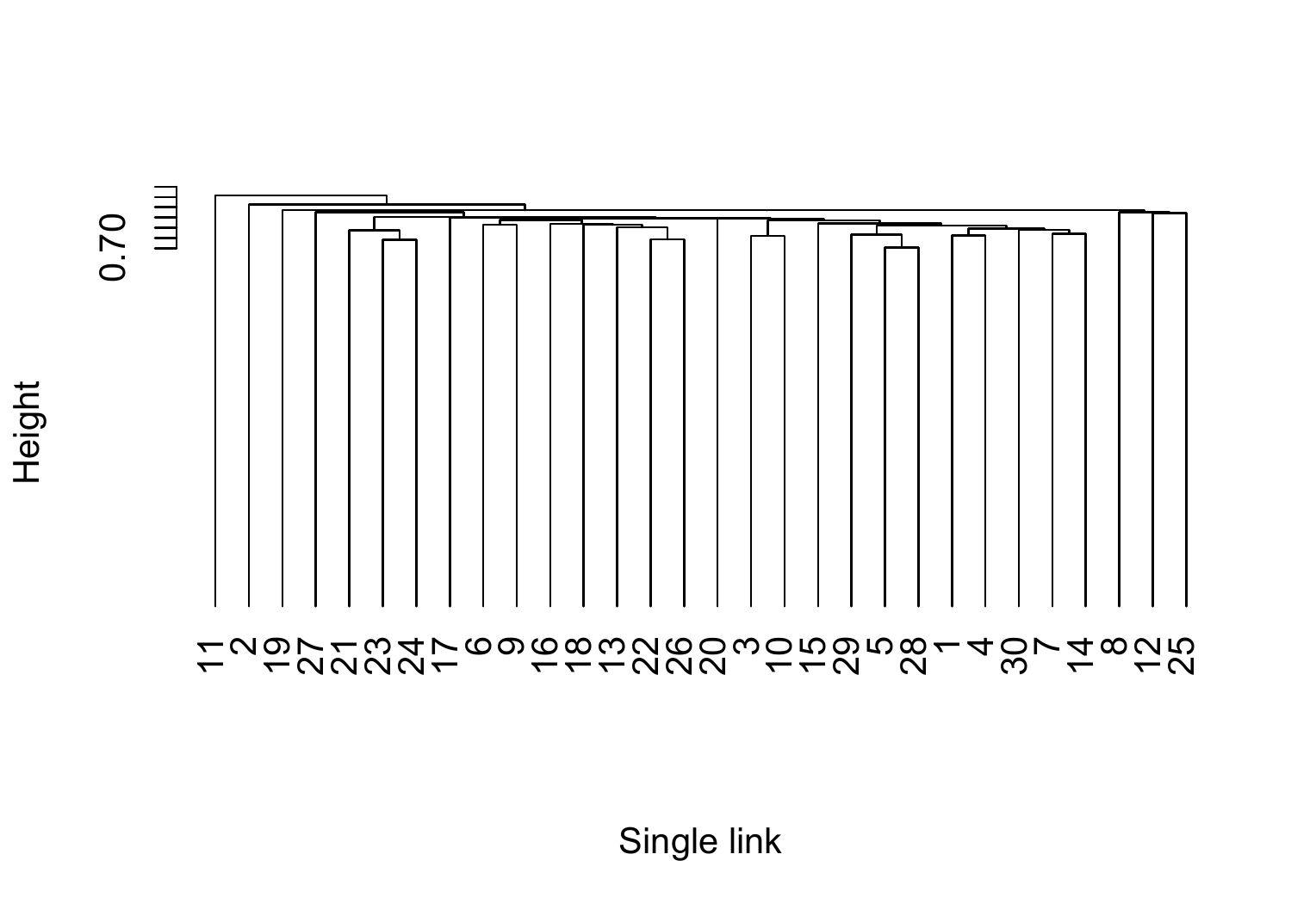}
\end{center}
\caption{Hierarchical clustering using the single link 
agglomerative criterion.  Input data are randomly generated.  
(See text for details.)} 
\label{fighran2}
\end{figure}

\begin{figure}
\begin{center}
\includegraphics[width=14cm]{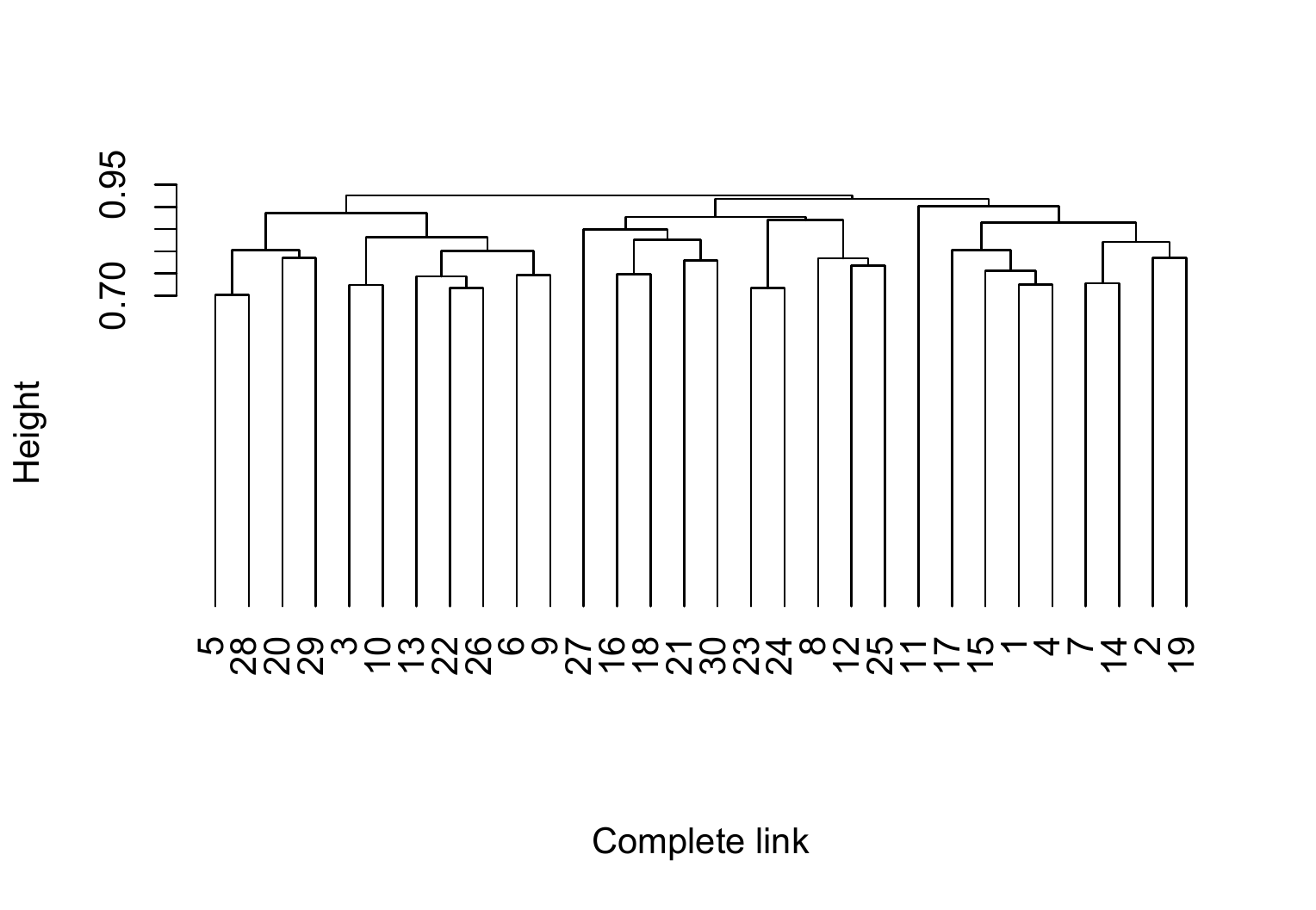}
\end{center}
\caption{Hierarchical clustering using the complete link 
agglomerative criterion.  Input data are randomly generated.  
(See text for details.)} 
\label{fighran3}
\end{figure}

We then induced hierarchies, with the agglomerative 
criteria noted -- Ward, Average, Single, Complete, McQuitty. 
Figures \ref{fighran1}, \ref{fighran2} and \ref{fighran3} show 
three of these hierarchies.

Table \ref{tabcons} has, throughout, higher values of ultrametric consistency 
between hierarchies, compared to Table \ref{tabconsran}.  While this is 
fully 
consistent in turn with greater ultrametricity in the initial data, which 
we have demonstrated to be so, nonetheless we issue the caveat here that 
what we are dealing with are hierarchies that are induced on the data.
Hence we need to be careful in drawing too many conclusions from this, 
insofar as the inducing of hierarchical relationship are forcing an 
increase in ultrametricity.  See also section \ref{seceps} below.

\subsection{From the Ultrametric Consensus Set to a Consensus Hierarchy}

We form a consensus hierarchy as follows.  For the two hierarchies, with 
ultrametric (cophenetic) distances that are denoted $d_{u1}$ 
and $d_{u2}$, we form a new set of ultrametric distances, 
denoted $d^{\mbox{Cons}}_{u}$, as follows.  As above, a morphologically 
consistent isosceles-with-small-base triplet means that the apex vertex 
has the same label.  

\begin{enumerate}
\item 
For all isosceles-with-small-base triples that are morphologically 
consistent, use the minimum, between the two hierarchies considered,
 pairwise distances:
$$d^{\mbox{Cons}}_u (i,j) \leftarrow \mbox{min} \{ 
d_{u1} (i,j) , d_{u2} (i,j) \} $$
\item For morphologically inconsistent triplets, we put all 
$$d^{\mbox{Cons}}_u (i,j), d^{\mbox{Cons}}_u (i,k), d^{\mbox{Cons}}_u (j,j)$$ 
in the triplet considered to be equal to the minimum of all triplet pair 
distances, i.e.\ the single minimum value,

$$ \mbox{min}
\{ d_{u1} (i,j), d_{u1} (i,k), d_{u1} (j,k), 
d_{u2} (i,j), d_{u2} (i,k), d_{u2} (j,k) \}$$
\end{enumerate}

We use the minimum in view of the maximal inferior ultrametric 
properties that ensue, i.e.\ the optimal fit from below
(cf.\ section \ref{msmi}).  

\begin{figure}
\begin{center}
\includegraphics[width=14cm]{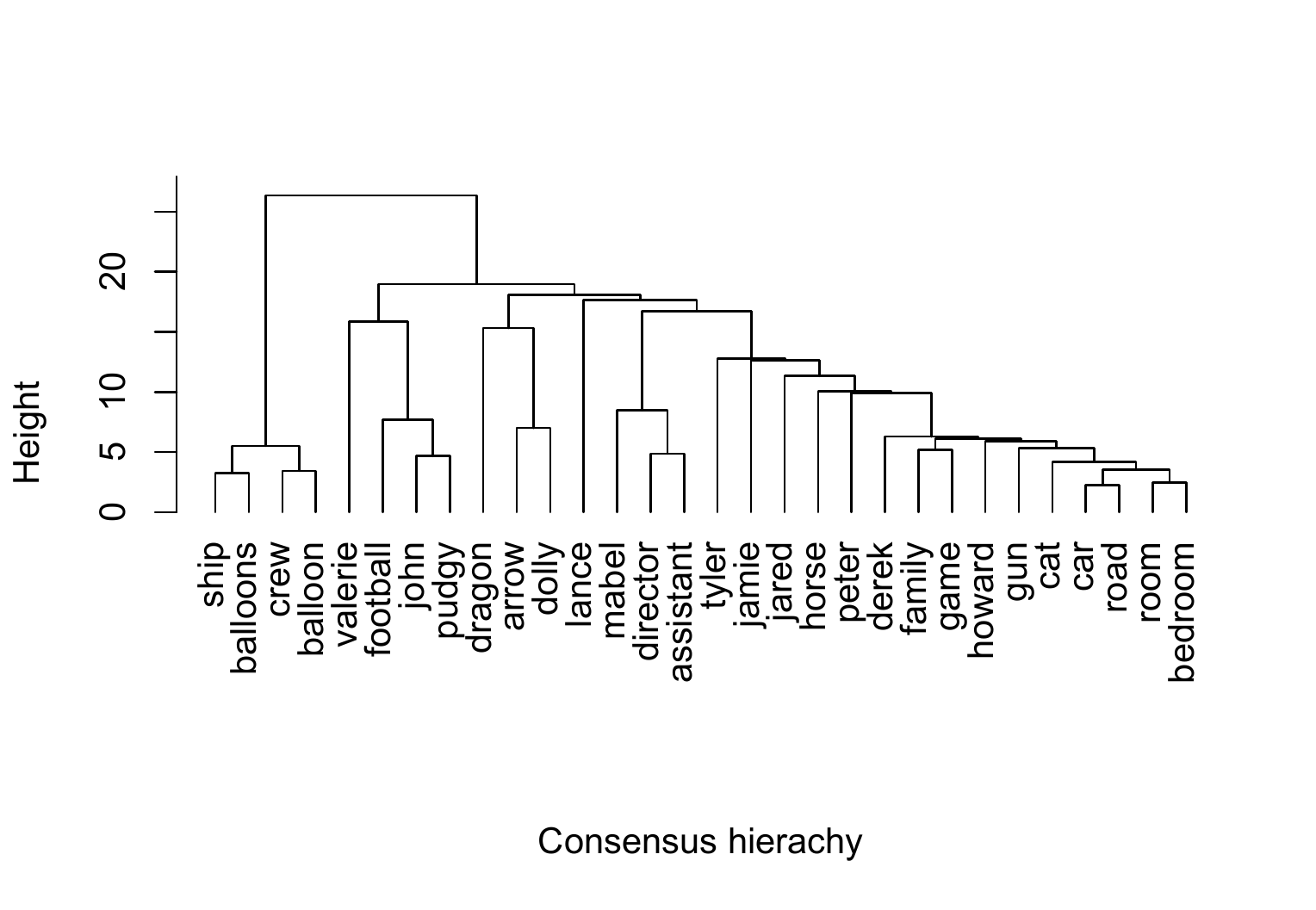}
\end{center}
\caption{Consensus hierarchical clustering using the Ward minimum variance 
and single link agglomerative criteria, with the Barbara Sanders 
data.  Hence consensus of Figures \ref{fig3hc1} and \ref{fig3hc2}.}
\label{hcons1}
\end{figure}

\begin{figure}
\begin{center}
\includegraphics[width=14cm]{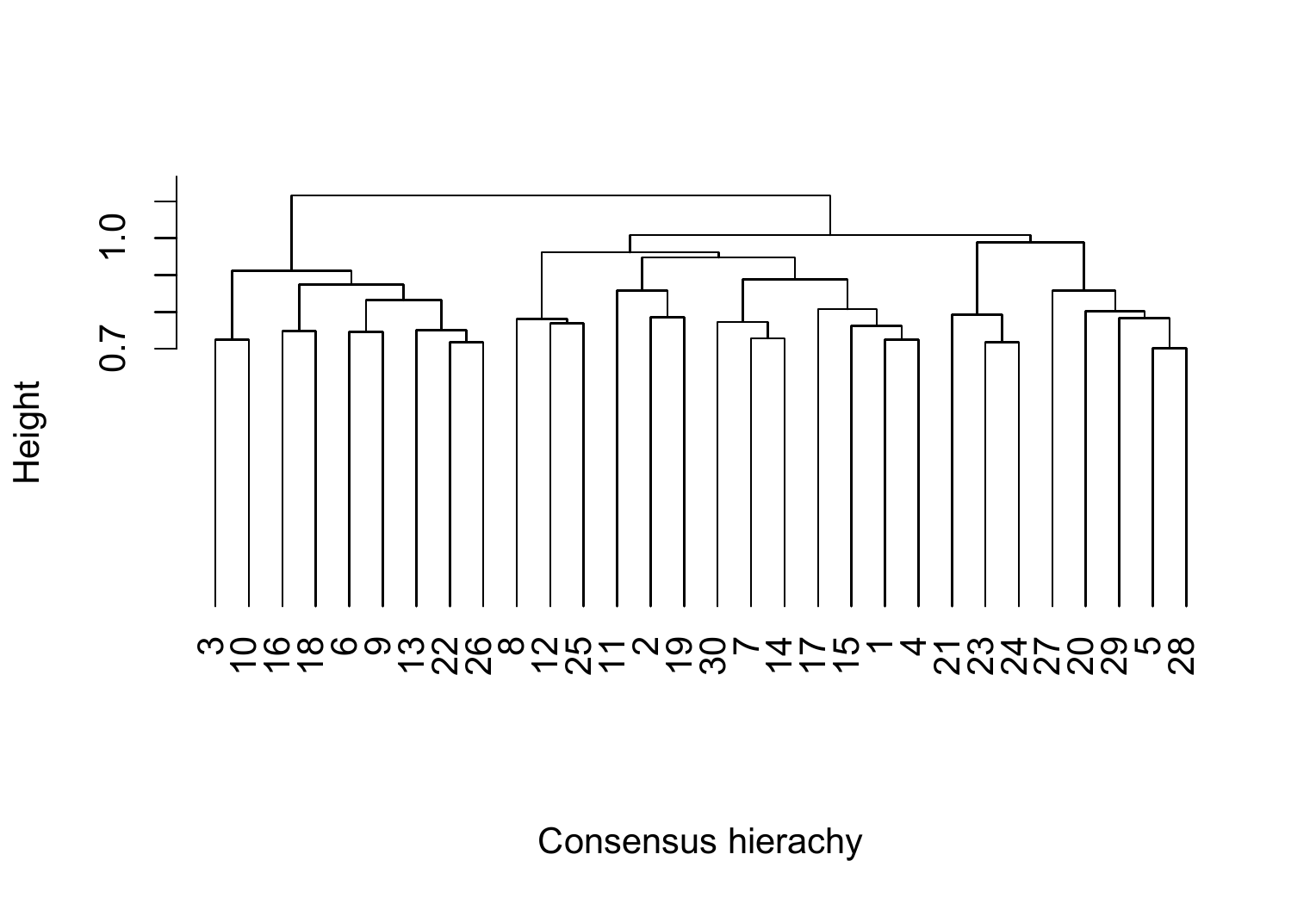}
\end{center}
\caption{Consensus hierarchical clustering using the Ward minimum variance 
and single link agglomerative criteria.  Random data used as input. 
Hence consensus of Figures \ref{fighran1} and \ref{fighran2}.}
\label{hcons2}
\end{figure}

The consensus hierarchy is such that the new hierarchy determined 
from the hierarchies associated with 
$u1$ and $u2$ is the same as the new hierarchy determined from the 
hierarchies associated with $u2$ and $u1$.  It is commutative and unique 
for a given input pair of hierarchies.   
 
A less good property of this consensus hierarchy is that it is a 
function of the pair of hierarchies used as input.  A different pair of 
input hierarchies will yield a somewhat different outcome.  

\section{Taking the Ultrametric Consensus Set Beyond Model Fitting: 
Inherent Ultrametric Properties}
\label{seceps}

\subsection{Coming from a Space Endowed at least with a Dissimilarity,
and Analyzing it when Mapped into an Ultrametric Space}

Thus we have a way to quantify ultrametric consensus, and the consensus set
in terms of triplets of observations.  Our next concern though is to 
recognize that the hierarchical clustering may be a good ultrametric embedding 
of our data.  But it is induced on our data.  More informally we can say
that it is forced on our data.  

Now we will relate ultrametric consensus 
back to our original, empirical and known data.

\subsection{Ultrametric Triplet Sets Through Other Pairs of Hierarchies}

A further reason to relate ultrametric consensus back to our 
original, empirical and known data is as follows.  

In the foregoing we looked at Ward and single link as the pair of hierarchies
from which we looked for consensus ultrametric triplets, and then proceeded
to find a subset of those triplets, 163 of them, 
 that were $\alpha_\epsilon$-ultrametric
in the original metric data space.  

Looking at the single and complete link hierarchies gave 190 
$\alpha_\epsilon$-ultrametric triplets.  Looking at Ward and complete link
gave 193 $\alpha_\epsilon$-ultrametric triplets.

We recall that any such consensus hierarchy provides us with the 
candidate set of $\alpha_\epsilon$-ultrametric parts of our data, 
for a given value of $\epsilon$.  We can easily determine the most
ultrametric parts of our data.  This we do from a consensus hierarchy  
pair in Appendix 3.  

\subsection{Just How Good Are the Ultrametric-Respecting Triplets 
Relative to the Input Data?}

\begin{figure}
\begin{center}
\includegraphics[width=14cm]{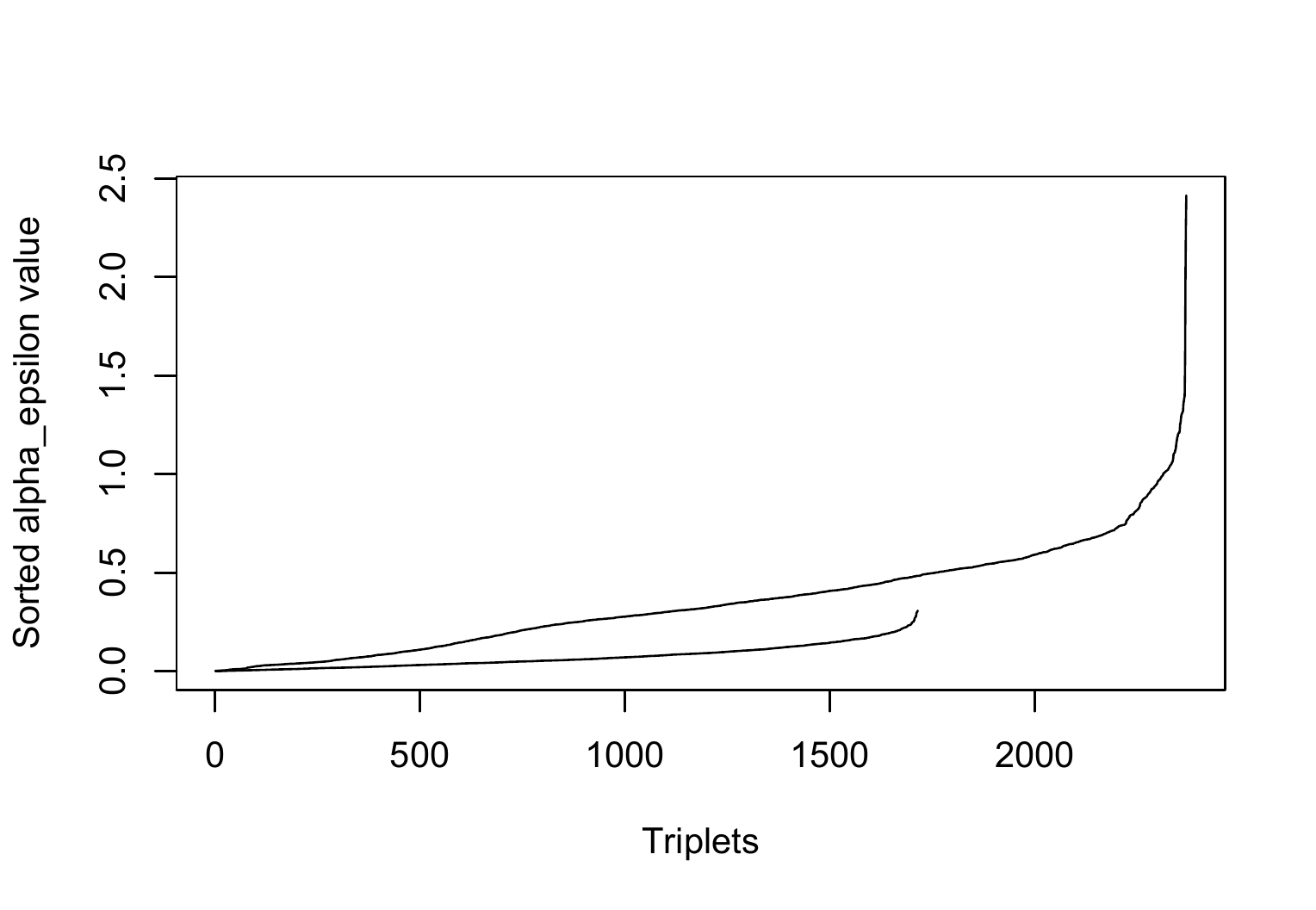}
\end{center}
\caption{[CORRECTION: ORDINATE IS EPSILON VALUES, IN RADIANS.  SEE NEXT FIGURE.]
The upper curve relates to the consensus of the Ward and 
single link hierarchies, see Figures \ref{fig3hc1} and \ref{fig3hc2}.  
The lower curve relates to the consensus of the Ward and single 
link hierarchies from random data, see Figures \ref{fighran1} and
\ref{fighran2}.  From Tables \ref{tabcons} and \ref{tabconsran}, 
we note that there were, respectively, 2369 and 1714 ultrametric-respecting
triplets.  The figure shows the sorted values of the $\alpha_\epsilon$ 
measure, determined from the input (i.e.\ to the hierarchical clustering)
data for the triplets.}  
\label{alphaepsplot}
\end{figure}

\begin{figure}
\begin{center}
\includegraphics[width=14cm]{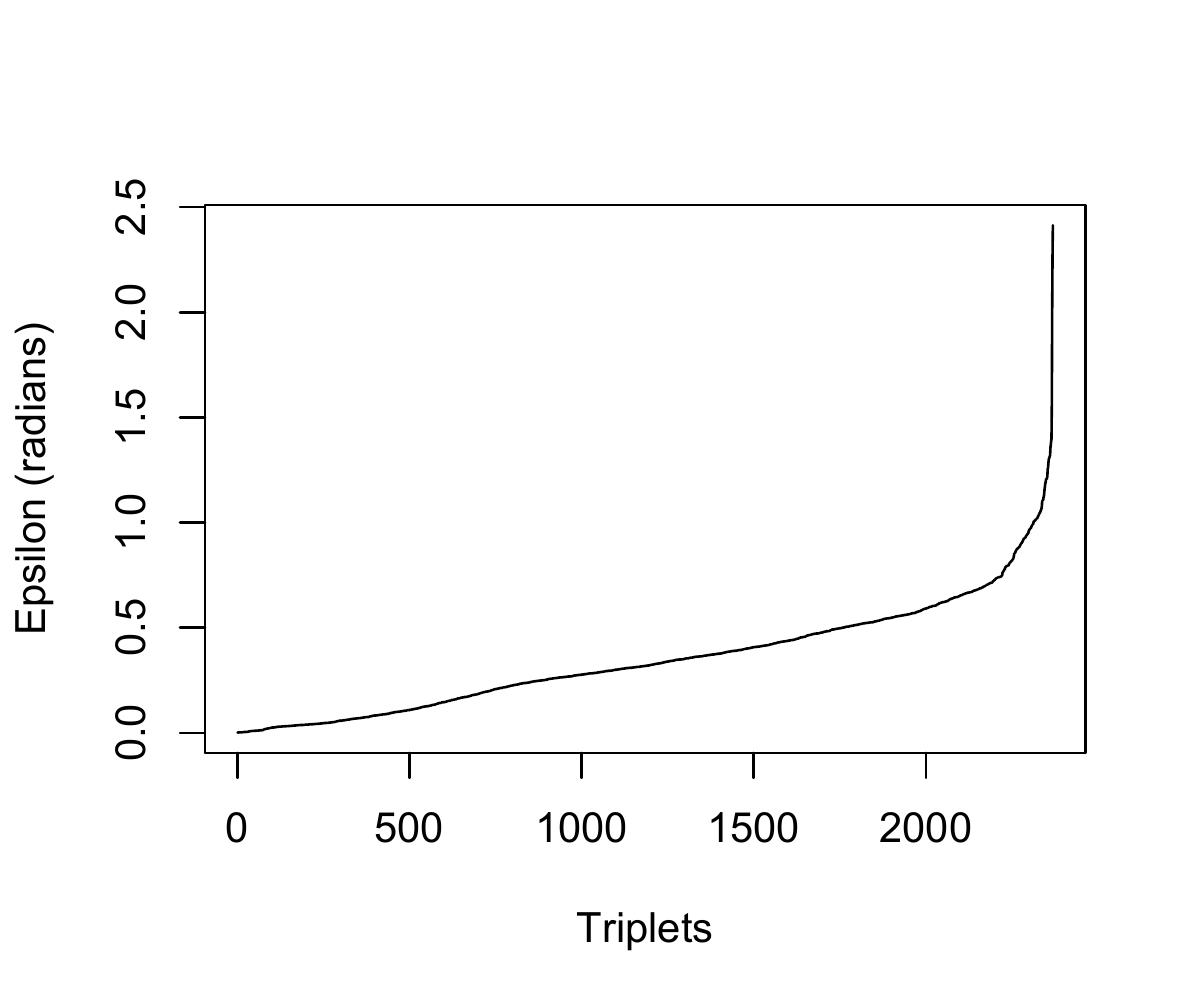}
\end{center}
\caption{CORRECTED VERSION OF PREVIOUS FIGURE -- LABEL ON ORDINATE.
From the consensus of two hierarchies (Ward and
single link hierarchies, see Figures \ref{fig3hc1} and \ref{fig3hc2})
all triplets are, by construction, ultrametric.  However, with reference
to their origin, as points in a Euclidean space (30 points in 
a 139-dimensional space), how close were they to being inherently 
ultrametric?  I.e.\ with very small $\epsilon$.  We see in the figure 
that very few are genuinely ultrametric, in this sense.  Typically we
use $\epsilon = 0.034906585$.}
\label{alphaepsplot2}
\end{figure}

Figure \ref{alphaepsplot} shows the values of $\alpha_\epsilon$. 
In previous work \cite{murtagh04,murtaghEPJ,murtaghmind2}, 
we took $\epsilon = 2$ degrees, or 0.034906585 radians.

Arising out of this look at a continuum of $\epsilon$ values, 
we conclude that neither the very large $\alpha_\epsilon$ values, 
nor any
other consideration, make us alter our view that $\epsilon = 2$ degrees, or 
0.034906585 radians is a satisfactory compromise between a small 
difference from identity ($\epsilon = 0$), while allowing a 
certain number of triangles to have this property of sufficiently 
nearly identical base angles.  

For the real data, we find 163 triplets, that are, as we know, 
consistent in isosceles-with-small-base morphology across two 
hierarchies (Ward, single link used), and that have an $\alpha_\epsilon$
value (i.e., difference in angles at the triplet triangle base)
$\leq 0.034906585$ radians.   For the random data, we find 549 
such triplets.  These numbers correspond to a threshold in Figure
\ref{alphaepsplot} of (ordinate) $0.034906585$ radians. 


Such triplets therefore have the following properties.

\begin{enumerate}
\item Each such triplet is embedded in an ultrametric space.  This 
implies that the associated triangle, in metric space terms, has a 
small base, and is isosceles.  
\item An amendment needs to be made to the foregoing point (as we have
noted before): we can
have equilateral triangles. The likelihood is small if (as in this 
work) we embed our data in a Euclidean space.  Our assertion here is 
that the number of identical distances on input is zero.  
\item By taking two hierarchical clusterings, each of which impose
an ultrametric embedding on the data, we determine a candidate set 
of consistent ultrametric triplet properties between the two 
hierarchies.  
\item A hierarchy, as noted, imposes an ultrametric structure on a
set of data.  Furthermore each such hierarchy, using as it does an
agglomerative clustering criterion, imposes a somewhat different 
hierarchical or ultrametric structure on the data.  This is so because
each hierarchy can be said to be approximating the data, and the criteria
are different (albeit based on relative compactness or density or linkage).  
\item We chose to work with the Ward minimum variance, and single link,
agglomerative clustering criteria because they were found to be more
distinct and contrasting.  The minimum variance criterion seeks good
compactness for the clusters formed (embedded in the hierarchy), while
the single link criterion is based on minimal connecting edges between
clusters.
\item  We have a consistent set of ultrametric triplets coming from
the two hierarchies.  In this candidate set of ultrametric triplets, 
and given that the hierarchies were ``imposed'' on the data (even if
in a particular best fit way), we ask how good the ultrametric 
properties are on the original input data,
\item Taking all candidate sets of ultrametric triplets, we determine
their $\alpha_\epsilon$ values for $\epsilon = 2$ degrees.  
\item $\alpha_\epsilon$ values are determined from the input data, 
i.e.\ the Euclidean-embedded, Correspondence Analysis output, that is 
full dimensional. 
\end{enumerate} 

\subsection{Interpretation of Ultrametric Triplets}

The figures to follow use the same planar projection as was used in 
Figure \ref{figca}.  The inherent dimensionality is 138, and
this was the basis
for the $\alpha_\epsilon$ property testing, and before that the 
metric context for inducing the hierarchical clustering. 

\begin{figure}
\begin{center}
\includegraphics[width=14cm]{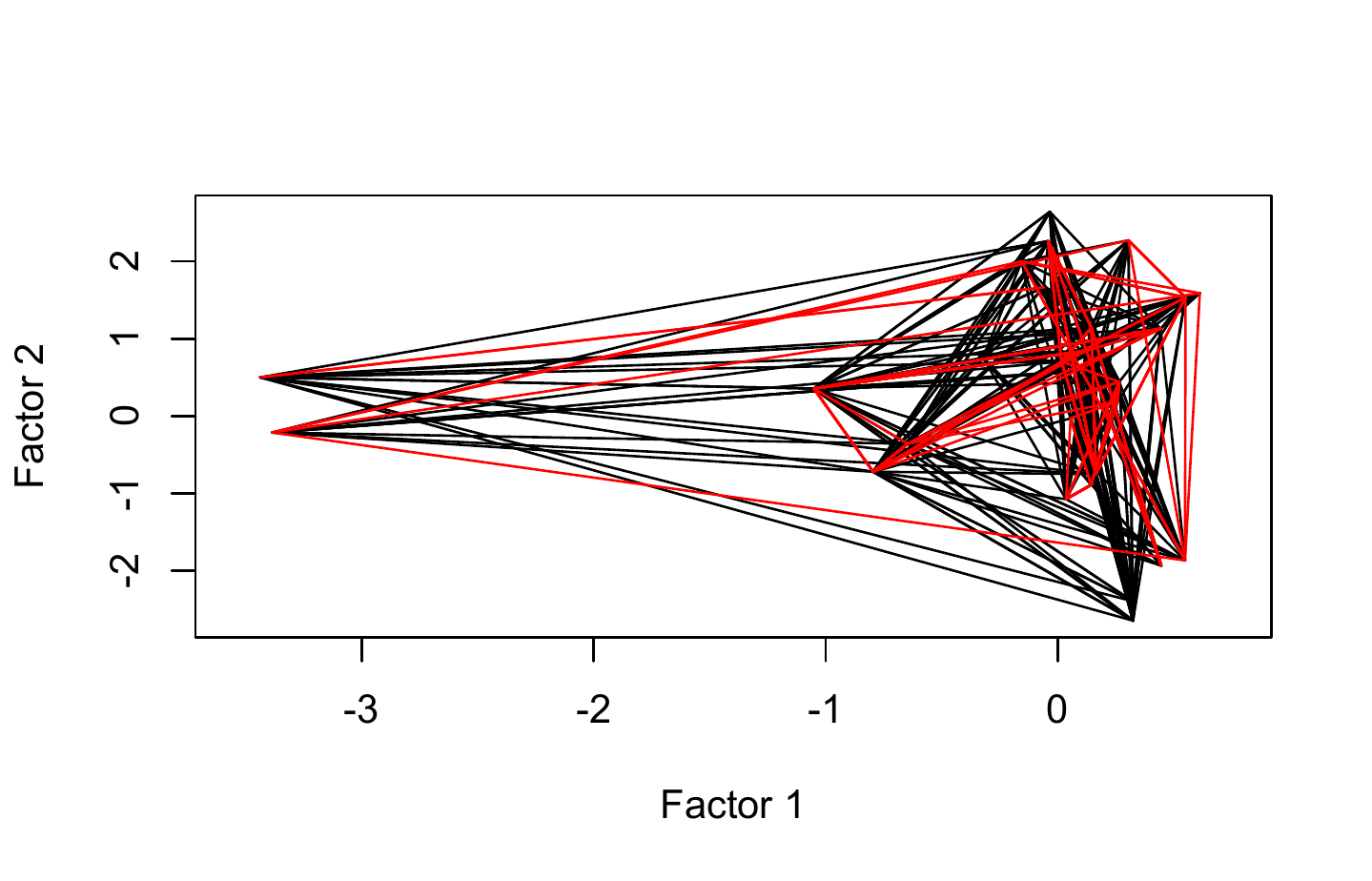}
\end{center}
\caption{The set of 163 ultrametric triplets,
coming from the consensus Ward and single link hierarchies, and 
respecting the $\alpha_\epsilon$ property for $\epsilon = 2$ degrees.
Edges drawn, the small base in the isoceles triangle shown in red.
Planar projection of the 138-dimensional space.  Cf.\ the same planar
projection in Figure \ref{figca}.}
\label{figall}
\end{figure}

In Figure \ref{figall}, the 163 triplets are depicted by a line 
segment connecting the three sides of each triangle.  The small 
base side is in red.  Recall that this is a depiction of isosceles
with small base triangles that satisfy the $\alpha_\epsilon$ property.
They are depicted here projected from their 138-dimensional space 
into the principal factor plane. 

We can look at detail in Figure \ref{figall} by taking terms separately.

\begin{figure}
\begin{center}
\includegraphics[width=14cm]{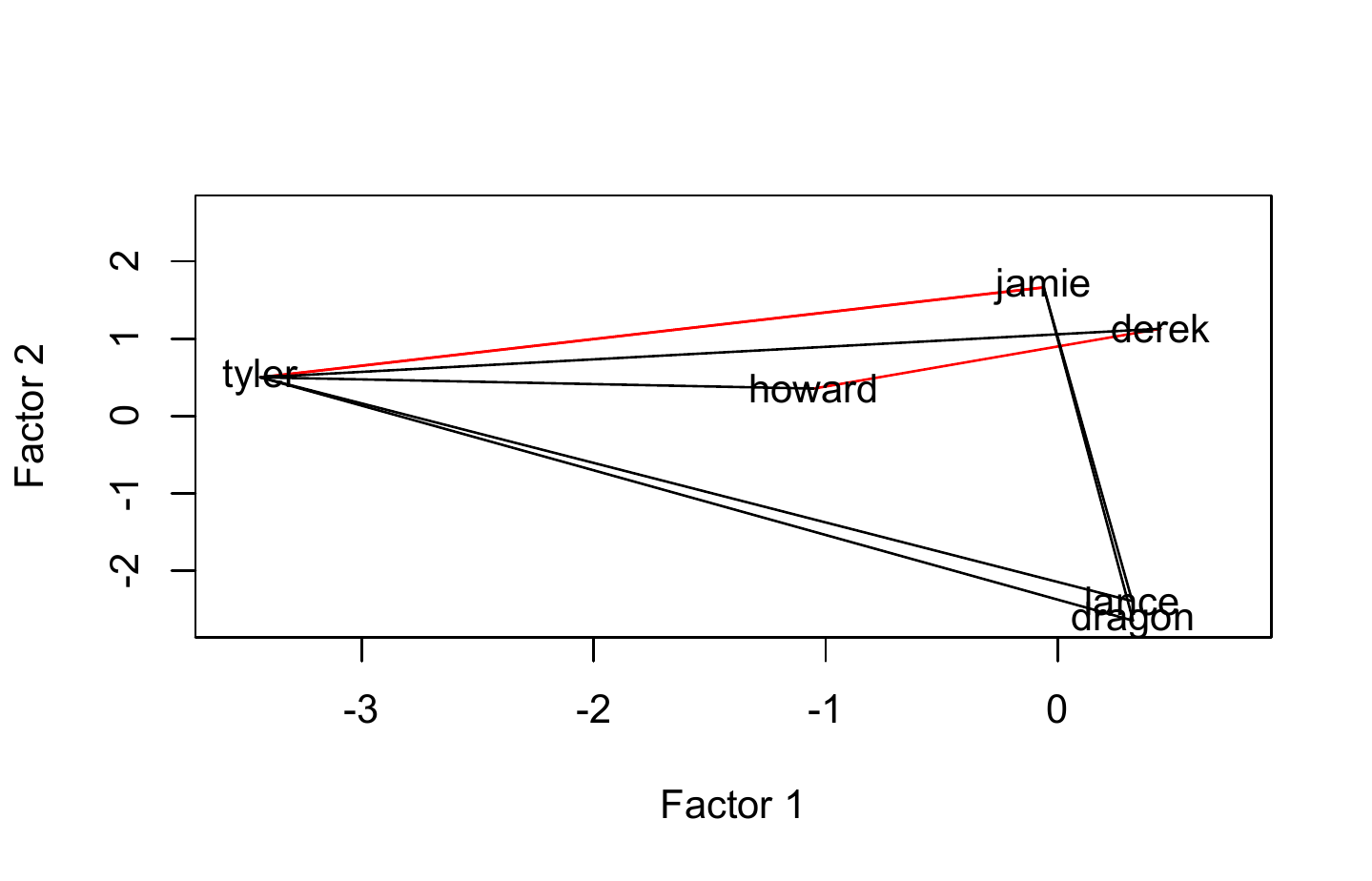}
\end{center}
\caption{The first three of the set of 163 ultrametric triplets,
coming from the consensus Ward and single link hierarchies, and 
respecting the $\alpha_\epsilon$ property for $\epsilon = 2$ degrees.
Edges drawn, the small base in the isoceles triangle shown in red.
Planar projection of the 138-dimensional space.  Cf.\ the same planar
projection in Figure \ref{figca}.}
\label{figfirst3}
\end{figure}

Figure \ref{figfirst3} depicts three ultrametric triplets 
respecting the $\alpha_\epsilon$ property for $\epsilon = 2$ degrees.
(To recap: the angles at the small base of the isosceles must be different 
by less than or equal to 2 degrees).  In terms of the 30 selected terms 
on which we are working, these triplets are as follows: 5, 29, 1; 1, 7, 10; 
and 1, 7, 12.  The small base is formed by the first two of the terms.  

The terms of the three ultrametric triplets are  as follows: 
1 = ``tyler''; 5 = ``derek''; 7 = ``jamie''; 
10 = ``dragon''; 12 = ``lance''; and 29 = ``howard''.
In Figure \ref{figfirst3}, the small base in the isosceles triangles
is in red.  For example, therefore, ``howard'' and ``derek'' form a 
small base, with triangle apex ``tyler''.   The names ``howard'' and ``derek''
are clustered closely relative to the more different name, ``tyler''.   
Similarly the close pair, ``jamie'' and ``tyler'' are contrasted both 
with the name ``lance'' and the term ``dragon''.   

The dates of the 139 Barbara Sanders dreams used were from 09/29/80 
(29 September 1980) to 01/25/97 (25 January 1997). 

According to \cite{bsanders}, personnages in the Barbara Sanders 
dreams are: 

\begin{itemize}
\item Howard: her ex-husband, divorced, died suddenly in 1997.
\item Derek: a man she had an affair with from 1994, and broke with in 1996.
\item Darryl: ex-boyfriend.
\item Dwight: favourite brother.
\end{itemize}

According to a look at the dream reports, this following are other
people and things: 

\begin{itemize}
\item Tyler (1980), ``co-worker''
\item Jamie (1981), ``old friend'', ``homosexual''
\item Mabel (1981), community college friend.   1991, ``co-worker''.
\item Peter (1980), ``a para, a student at the community college'', married.
\item horse (1981 to 1991, but mostly in a 1980 and a 1981 dream).
\item dragon (1993) (no other names mentioned in this dream about a dragon)
\item Lance (1992) ``is black and used to be the city manager assistant
  to the disability rights city group''; ``married''.
\item game, gun, family: words used (in differing contexts) throughout the years.

\end{itemize}

The full 163 triplet set of ultrametric cases is in Appendix A3.  The 
following is a small extracted set of these.  Note that labels are 
alphabetically ordered for the two base vertices of the triangle, columns
1 and 2, and the triangle apex is in column 3.  Then comes the 
$\alpha_\epsilon$ difference of base angles, in radians.  

Note, in the following, the later period (i.e.\ 1992, 1993) dragon, Lance, 
counterposed to the earlier period (i.e. 1980, 1981) presence
of Tyler, Jamie, Mabel, Peter, horse.  

\begin{verbatim}
  [9,] "tyler"     "jamie"     "dragon"    "0.002900765"
 [10,] "tyler"     "jamie"     "lance"     "0.002900765"

 [47,] "family"    "gun"       "jamie"     "0.01004061"
 [48,] "family"    "gun"       "tyler"     "0.01007054"

 [68,] "peter"     "horse"     "jamie"     "0.01281991"
 [69,] "peter"     "horse"     "tyler"     "0.012843"

 [40,] "peter"     "mabel"     "dragon"    "0.009091976"
 [41,] "peter"     "mabel"     "lance"     "0.009091976"
 [52,] "peter"     "horse"     "dragon"    "0.01065118"
 [53,] "peter"     "horse"     "lance"     "0.01065118"
 [81,] "mabel"     "horse"     "dragon"    "0.01887733"
 [82,] "mabel"     "horse"     "lance"     "0.01887733"

 [26,] "game"      "howard"    "jamie"     "0.005031676"
 [27,] "game"      "howard"    "tyler"     "0.005045841"

 [16,] "game"      "howard"    "dragon"    "0.003874804"
 [17,] "game"      "howard"    "lance"     "0.003874804"
[102,] "derek"     "howard"    "dragon"    "0.02580133"
[103,] "derek"     "howard"    "lance"     "0.02580133"
[134,] "derek"     "game"      "dragon"    "0.03133084"
[135,] "derek"     "game"      "lance"     "0.03133084"
\end{verbatim}

At least as regards this selection, we observe 
association of:

\begin{itemize} 
\item Tyler, Jamie; 
\item family, gun; 
\item Peter, horse; Mabel, horse; Peter, Mabel; 
\item Howard, game; Derek, game; Howard, Derek.
\end{itemize}

We also see close association, in this selection, between:

\begin{itemize}
\item Tyler, Jamie;
\item Howard, Derek.
\end{itemize}

In this short exploration, we see how we can focus in on 
particular associations of names or of terms, and also on 
contrasting names and terms.  Through ultrametricity we have
a ranking from near perfect ($\alpha_\epsilon = 0$) up to 
our imposed limit of a 2 degree difference between base 
angles in triangle configurations.  

\section{Conclusion}

Our analysis of triangle properties can be related to triads, or
triadic patterns, in other areas of speech or writing.  Veale 
\cite{veale} notes the following: 
``Recent work by cognitive scientists Jeffrey Loewenstein and Chip 
Heath 
shows that the AAB pattern in stories -- which they call the 
repetition-break plot structure -- is considered more enjoyable 
by readers than the equivalent AAA (unbroken repetition) or
ABC (no repetition) patterns. Many narrative jokes use explicit 
repetition to enforce an AAA pattern in the minds of an audience, 
so that AAB repetition-break comes as an incongruous and 
potentially humorous surprise.''

He cites \cite{lowenstein} and \cite{rozin}.  He gives various 
examples, and adds: 
``There are whole genres of jokes involving a priest, a rabbi and an imam; 
or an Irishman, and Englishman and a Scotsman; or a trio of nuns, hookers, 
husbands or some other stock characters, in which two of the three act 
somewhat predictably while the zany actions of the third provide the 
humorous departure.''

The above is noted as another example of where an ultrametric 
triangle provides a visualization for the two base vertices of a 
triangle, followed by a quite distinct, and possibly distant, triangle
apex.  

Future work will aim at addressing 
humour and jokes, through use
of an encoding of successive assertions (rather that the word-based 
approach that was developed in this work described in this paper).   

\section*{Appendix}
\subsection*{A1. Comparing and Combining Hierarchical Clusterings}

The ultrametric distance associated with a dendrogram 
(see section \ref{sectUMdist}) is commonly termed 
cophenetic distance \cite{sokal}.  This distance is the smallest
one for which two observations are in the same cluster.  Informally
expressed it is the distance to the closest common ancestor.
Comparing this distance
with the input distances used leads to a correlation measure, cophenetic
correlation \cite{rohlf}.   The latter is a goodness of fit
measure between input and the hierarchy that is fit to the data.

By comparing the cophenetic distance matrices we also have a means of 
assessing how close two dendrograms are to one another.  Hence this is the
most immediate and most widely used approach to comparison of hierarchical
clusterings.  See \cite{legendre} for discussion and applications.  

Given two or more hierarchical clusterings, 
Zheng et al.\ \cite{zheng} combine ultrametric distances in a comparative
study involving (i) various ultrametric distances derived from a
dendrogram, including cophenetic distance; rank distance; and cluster,
partition and subtree cardinality measures; (ii) adding ultrametric
distance matrices over the set of dendrograms, and then using the
transitive dissimilarity (to be explained next) as a means to force
this aggregated set of ultrametrics to be, itself, ultrametric.

On a path connecting vertices $i$ and $j$, consider the transitive
dissimilarity as the maximum edge length on that path.  Then over all
paths between the two vertices, the minimal transitive dissimilarity is
determined.  This is an ultrametric.   (A modified Floyd-Warshall
transitive closure algorithm is used to determine these new distances
between all $i$ and $j$, in \cite{zheng}).

Path and subtree combinatorial optimization represents a commonly
used approach for tree difference or modification, e.g.\ \cite{farach} 
or \cite{lee}. 

Morlini and Zani \cite{morlini} define and study in depth a new dissimilarity
measure between dendrograms that takes into account the partitions
and clusters, as well as the embeddedness relationships.

\subsection*{A2. Metrics and Ultrametrics based on 3-Way Distances}

Our $\alpha_\epsilon$-ultrametricity measure involves looking
at, and testing for ultrametricity, in triangles.  Each triangle must be 
isosceles-with-small-base, or equilateral, to be a case of an ultrametric
relation in the triplet considered.  The measure $\alpha_\epsilon$ considers
a triplet $i, j, k$ and looks at distances $d_{ij}, d_{ik}, d_{jk}$ or 
angles between them. 

An idea is to take the consideration of ultrametric or not, using the 
triplet $i, j, k$, to the different context of a 3-way distance, $d_{ijk}$.  
In \cite{joly}, such 3-way distances are defined, and termed: semi-perimeter
distance, star distance, inertial distance, and restriction of 3-way 
distances to 2-way.  Three-way incidence tables, i.e.\ tensors, are 
considered, and the minimal spanning tree is generalized to triplets 
where ``connecting edges by an extremal point'' becomes instead 
``connecting triangles by a common side''.   

The matching of triplet index matrices (tensors) and fitting a tree under 
the $\ell_\infty$ or max norm are considered in \cite{gronau}. 

\subsection*{A3.  Ultrametric Alpha Epsilon-Respecting Triplets 
Using Ward and Single Link}

There are 163 $\alpha_\epsilon$ property triplets in the following. The 2 degrees
capped difference between the base angles (of the isosceles-with-small-base
triangles) equals 0.034906585 radians.  Column 4 in the following
gives the actual difference of angles.  The other columns are such
that column 3 is the triangle apex.   

{\small
\begin{verbatim}
       [,1]        [,2]        [,3]        [,4]         
  [1,] "family"    "gun"       "arrow"     "0.001851383"
  [2,] "car"       "road"      "cat"       "0.002350057"
  [3,] "road"      "bedroom"   "balloon"   "0.002403404"
  [4,] "assistant" "horse"     "dragon"    "0.002445828"
  [5,] "assistant" "horse"     "valerie"   "0.002493735"
  [6,] "assistant" "horse"     "ship"      "0.002621979"
  [7,] "assistant" "horse"     "balloons"  "0.002705753"
  [8,] "assistant" "horse"     "balloon"   "0.002866144"
  [9,] "tyler"     "jamie"     "dragon"    "0.002900765"
 [10,] "tyler"     "jamie"     "lance"     "0.002900765"
 [11,] "john"      "director"  "balloon"   "0.002964512"
 [12,] "john"      "dolly"     "balloon"   "0.002971291"
 [13,] "bedroom"   "cat"       "pudgy"     "0.003025777"
 [14,] "assistant" "horse"     "crew"      "0.003123132"
 [15,] "car"       "bedroom"   "cat"       "0.003820139"
 [16,] "game"      "howard"    "dragon"    "0.003874804"
 [17,] "game"      "howard"    "lance"     "0.003874804"
 [18,] "family"    "assistant" "lance"     "0.003954736"
 [19,] "game"      "howard"    "ship"      "0.004253321"
 [20,] "game"      "howard"    "balloons"  "0.004444716"
 [21,] "game"      "howard"    "director"  "0.004576373"
 [22,] "game"      "howard"    "dolly"     "0.004598292"
 [23,] "road"      "bedroom"   "cat"       "0.004736657"
 [24,] "game"      "howard"    "balloon"   "0.004836543"
 [25,] "director"  "dolly"     "valerie"   "0.004974686"
 [26,] "game"      "howard"    "jamie"     "0.005031676"
 [27,] "game"      "howard"    "tyler"     "0.005045841"
 [28,] "director"  "dolly"     "ship"      "0.00514599" 
 [29,] "car"       "road"      "game"      "0.005168709"
 [30,] "director"  "dolly"     "balloons"  "0.005252192"
 [31,] "game"      "howard"    "crew"      "0.005558694"
 [32,] "game"      "howard"    "arrow"     "0.006075735"
 [33,] "mabel"     "assistant" "game"      "0.006910084"
 [34,] "family"    "gun"       "dragon"    "0.007645213"
 [35,] "mabel"     "assistant" "road"      "0.007757971"
 [36,] "family"    "gun"       "valerie"   "0.007849058"
 [37,] "family"    "gun"       "ship"      "0.008419404"
 [38,] "family"    "gun"       "balloons"  "0.008814132"
 [39,] "family"    "gun"       "director"  "0.009087052"
 [40,] "peter"     "mabel"     "dragon"    "0.009091976"
 [41,] "peter"     "mabel"     "lance"     "0.009091976"
 [42,] "family"    "gun"       "dolly"     "0.009132605"
 [43,] "family"    "gun"       "john"      "0.009603296"
 [44,] "family"    "gun"       "balloon"   "0.009629988"
 [45,] "peter"     "assistant" "dragon"    "0.009635763"
 [46,] "peter"     "mabel"     "ship"      "0.009780957"
 [47,] "family"    "gun"       "jamie"     "0.01004061" 
 [48,] "family"    "gun"       "tyler"     "0.01007054" 
 [49,] "peter"     "mabel"     "balloons"  "0.01011235" 
 [50,] "peter"     "assistant" "ship"      "0.0103711"  
 [51,] "bedroom"   "assistant" "lance"     "0.01064924" 
 [52,] "peter"     "horse"     "dragon"    "0.01065118" 
 [53,] "peter"     "horse"     "lance"     "0.01065118" 
 [54,] "peter"     "assistant" "balloons"  "0.01072554" 
 [55,] "peter"     "mabel"     "balloon"   "0.01075504" 
 [56,] "family"    "gun"       "jared"     "0.0108418"  
 [57,] "family"    "gun"       "football"  "0.01129312" 
 [58,] "football"  "pudgy"     "director"  "0.01135783" 
 [59,] "peter"     "assistant" "balloon"   "0.0114148"  
 [60,] "peter"     "horse"     "ship"      "0.01142248" 
 [61,] "family"    "gun"       "pudgy"     "0.01150559" 
 [62,] "peter"     "horse"     "balloons"  "0.01178966" 
 [63,] "peter"     "mabel"     "crew"      "0.01181512" 
 [64,] "family"    "gun"       "peter"     "0.01193978" 
 [65,] "football"  "dolly"     "crew"      "0.01230778" 
 [66,] "peter"     "horse"     "balloon"   "0.0124934"  
 [67,] "peter"     "assistant" "crew"      "0.0125593"  
 [68,] "peter"     "horse"     "jamie"     "0.01281991" 
 [69,] "peter"     "horse"     "tyler"     "0.012843"   
 [70,] "mabel"     "assistant" "family"    "0.0129161"  
 [71,] "mabel"     "assistant" "car"       "0.01301065" 
 [72,] "peter"     "horse"     "jared"     "0.01340517" 
 [73,] "jared"     "football"  "crew"      "0.01352421" 
 [74,] "peter"     "horse"     "crew"      "0.01362332" 
 [75,] "game"      "gun"       "assistant" "0.01414658" 
 [76,] "family"    "howard"    "peter"     "0.01492026" 
 [77,] "road"      "room"      "family"    "0.01493067" 
 [78,] "mabel"     "assistant" "cat"       "0.01692686" 
 [79,] "family"    "howard"    "mabel"     "0.01835927" 
 [80,] "car"       "room"      "crew"      "0.01850425" 
 [81,] "mabel"     "horse"     "dragon"    "0.01887733" 
 [82,] "mabel"     "horse"     "lance"     "0.01887733" 
 [83,] "mabel"     "horse"     "valerie"   "0.01925078" 
 [84,] "derek"     "gun"       "balloon"   "0.01943399" 
 [85,] "car"       "cat"       "mabel"     "0.019929"   
 [86,] "mabel"     "horse"     "ship"      "0.02025176" 
 [87,] "mabel"     "horse"     "balloons"  "0.02090672" 
 [88,] "arrow"     "mabel"     "valerie"   "0.021144"   
 [89,] "mabel"     "assistant" "bedroom"   "0.02164229" 
 [90,] "mabel"     "horse"     "balloon"   "0.02216332" 
 [91,] "room"      "cat"       "assistant" "0.0221949"  
 [92,] "arrow"     "mabel"     "ship"      "0.02227254" 
 [93,] "car"       "road"      "director"  "0.02242459" 
 [94,] "arrow"     "mabel"     "balloons"  "0.02301388" 
 [95,] "family"    "howard"    "horse"     "0.02331366" 
 [96,] "bedroom"   "cat"       "balloons"  "0.02342833" 
 [97,] "mabel"     "horse"     "crew"      "0.02418521" 
 [98,] "bedroom"   "family"    "lance"     "0.024372"   
 [99,] "arrow"     "mabel"     "balloon"   "0.0244434"  
[100,] "mabel"     "assistant" "dragon"    "0.02544971" 
[101,] "bedroom"   "cat"       "peter"     "0.02573532" 
[102,] "derek"     "howard"    "dragon"    "0.02580133" 
[103,] "derek"     "howard"    "lance"     "0.02580133" 
[104,] "mabel"     "assistant" "valerie"   "0.02600296" 
[105,] "car"       "bedroom"   "balloon"   "0.02604236" 
[106,] "derek"     "family"    "balloon"   "0.02616928" 
[107,] "derek"     "howard"    "valerie"   "0.02645994" 
[108,] "car"       "road"      "valerie"   "0.02659476" 
[109,] "arrow"     "mabel"     "crew"      "0.02676692" 
[110,] "car"       "road"      "ship"      "0.02690026" 
[111,] "pudgy"     "horse"     "ship"      "0.02716962" 
[112,] "mabel"     "assistant" "ship"      "0.02750808" 
[113,] "pudgy"     "horse"     "balloons"  "0.02801975" 
[114,] "derek"     "howard"    "ship"      "0.02828986" 
[115,] "mabel"     "assistant" "balloons"  "0.02851275" 
[116,] "mabel"     "assistant" "dolly"     "0.02930135" 
[117,] "gun"       "howard"    "peter"     "0.0293042"  
[118,] "road"      "bedroom"   "dolly"     "0.02939051" 
[119,] "derek"     "howard"    "balloons"  "0.02954464" 
[120,] "pudgy"     "horse"     "balloon"   "0.02964163" 
[121,] "assistant" "pudgy"     "ship"      "0.02978986" 
[122,] "room"      "cat"       "horse"     "0.03020312" 
[123,] "road"      "bedroom"   "peter"     "0.03026897" 
[124,] "car"       "road"      "balloons"  "0.03030422" 
[125,] "derek"     "howard"    "director"  "0.03040631" 
[126,] "mabel"     "assistant" "john"      "0.03043118" 
[127,] "car"       "road"      "football"  "0.03048117" 
[128,] "mabel"     "assistant" "balloon"   "0.03049399" 
[129,] "derek"     "howard"    "dolly"     "0.03054964" 
[130,] "assistant" "pudgy"     "balloons"  "0.03072369" 
[131,] "football"  "pudgy"     "dragon"    "0.03104475" 
[132,] "football"  "pudgy"     "lance"     "0.03104475" 
[133,] "peter"     "arrow"     "ship"      "0.03109828" 
[134,] "derek"     "game"      "dragon"    "0.03133084" 
[135,] "derek"     "game"      "lance"     "0.03133084" 
[136,] "mabel"     "assistant" "jamie"     "0.03144317" 
[137,] "mabel"     "assistant" "tyler"     "0.03151111" 
[138,] "family"    "howard"    "dragon"    "0.03191839" 
[139,] "jared"     "director"  "crew"      "0.03197288" 
[140,] "derek"     "howard"    "john"      "0.03202227" 
[141,] "jared"     "dolly"     "crew"      "0.03206637" 
[142,] "peter"     "arrow"     "balloons"  "0.03212724" 
[143,] "pudgy"     "horse"     "crew"      "0.03222202" 
[144,] "car"       "room"      "assistant" "0.03243193" 
[145,] "assistant" "pudgy"     "balloon"   "0.03250582" 
[146,] "director"  "pudgy"     "crew"      "0.03258847" 
[147,] "pudgy"     "dolly"     "crew"      "0.03268863" 
[148,] "family"    "howard"    "valerie"   "0.03275107" 
[149,] "jared"     "john"      "crew"      "0.03299317" 
[150,] "family"    "game"      "dragon"    "0.03302905" 
[151,] "cat"       "gun"       "game"      "0.03313567" 
[152,] "mabel"     "assistant" "jared"     "0.03320492" 
[153,] "derek"     "howard"    "jamie"     "0.03337614" 
[154,] "derek"     "howard"    "tyler"     "0.03346827" 
[155,] "football"  "pudgy"     "ship"      "0.03347461" 
[156,] "car"       "road"      "balloon"   "0.03358371" 
[157,] "mabel"     "assistant" "crew"      "0.0338856"  
[158,] "bedroom"   "gun"       "howard"    "0.03408125" 
[159,] "peter"     "arrow"     "balloon"   "0.03410934" 
[160,] "mabel"     "assistant" "football"  "0.03414623" 
[161,] "derek"     "game"      "ship"      "0.03438099" 
[162,] "mabel"     "assistant" "pudgy"     "0.03457698" 
[163,] "football"  "pudgy"     "balloons"  "0.03465437" 
\end{verbatim}
}

\bibliographystyle{plain}
\bibliography{biblio}

\end{document}